\documentclass[10pt,journal,cspaper,compsoc]{IEEEtran}
\usepackage{epsfig}
\usepackage{graphicx}
\usepackage{amsmath}
\usepackage{amssymb}
\usepackage{amsthm}
\usepackage{color,multirow}
\usepackage{xspace}
\usepackage[pagebackref=true,breaklinks=true,citecolor=black,linkcolor=red,letterpaper=true,colorlinks,bookmarks=false]{hyperref}

\usepackage{contour}
\usepackage{xkeyval}
\usepackage[dvipsnames]{xcolor}

\newcommand{\mv}[1]{\mathbf{#1}}
\def\eg{\emph{e.g.}\xspace} 
\def\ie{\emph{i.e.}\xspace} 
 
\def\etc{\emph{etc.}\xspace} 
 
\def\etal{\emph{et al.}\xspace}

\newcommand{\bd}[1]{\textbf{#1}}
\newcommand{\ud}[1]{#1}

\newcommand{\beforePara}{\vspace{-0em}}

\newcommand{\beforeEqn}{\vspace{-0em}}

\newcommand{\tablestyle}[2]{\setlength{\tabcolsep}{#1}\renewcommand{\arraystretch}{#2}\centering\footnotesize}

\usepackage{tabulary}
\usepackage[caption=false]{subfig}

\newcolumntype{x}[1]{>{\centering\arraybackslash}p{#1pt}}

\newcommand{\wh}{\color{white}} 
\contourlength{.5pt}
\newcommand{\figlabel}[1]{\sffamily\bfseries\wh\scriptsize\contour{black}{#1}} 
\makeatletter
\newlength{\sfp@hseplen}\newlength{\sfp@vseplen}
\define@cmdkey{subfigpos}[sfp@]{vsep}[0.6\baselineskip]{\setlength{\sfp@vseplen}{\sfp@vsep}}
\define@cmdkey{subfigpos}[sfp@]{hsep}[2.5pt]{\setlength{\sfp@hseplen}{\sfp@hsep}}
\newcommand{\subfigimg}[3][,]{%
	\setkeys{Gin,subfigpos}{vsep,hsep,#1}
	\setbox1=\hbox{\includegraphics{#3}}
	\leavevmode\rlap{\usebox1}
	\rlap{\hspace*{8pt}\raisebox{\dimexpr\ht1-9pt}{\figlabel{#2}}}
	\phantom{\usebox1}
}

\newcommand{\ignore}[1]{}
\ifCLASSOPTIONcompsoc
  \usepackage[nocompress]{cite}
\else
  \usepackage{cite}
\fi
\ifCLASSINFOpdf
\else
\fi

\begin{document}
%
\title{Models Matter, So Does Training: An Empirical Study of CNNs for Optical Flow Estimation}

\author{Deqing Sun, Xiaodong Yang, Ming-Yu Liu, and Jan Kautz
\IEEEcompsocitemizethanks{\IEEEcompsocthanksitem D. Sun and J. Kautz are with NVIDIA, Westford, MA 01886.\protect\\
E-mail: \{deqings, jkautz\}@nvidia.com
\IEEEcompsocthanksitem X. Yang and M.-Y. Liu are with NVIDIA, Santa Clara, CA 95050.\protect\\
E-mail: \{xiaodongy, mingyul\}@nvidia.com}
}

%
%


\IEEEcompsoctitleabstractindextext{	
    \begin{abstract}
    We investigate two crucial and closely related aspects of CNNs for optical flow estimation: models and training. 
	First, we design a compact but effective CNN model, called PWC-Net, according to simple and well-established principles: pyramidal processing, warping, and cost volume processing. 
	PWC-Net is 17 times smaller in size, 2 times faster in inference, and 11\% more accurate on Sintel final than the recent FlowNet2 model. It is the winning entry in the optical flow competition of the robust vision challenge.
	Next, we experimentally analyze the sources of our performance gains.
	In particular, we use the same training procedure of PWC-Net to retrain FlowNetC, a sub-network of FlowNet2. 
	The retrained FlowNetC is 56\% more accurate on Sintel final than the previously trained one and even 5\% more accurate than the FlowNet2 model.
	We further improve the training procedure and increase the accuracy of PWC-Net on Sintel by 10\% and on KITTI 2012 and 2015 by 20\%.
	Our newly trained model parameters and training protocols will be available on \url{https://github.com/NVlabs/PWC-Net}. 
	\end{abstract}
\begin{IEEEkeywords}
Optical flow, pyramid, warping, cost volume, and convolutional neural network (CNN).
\end{IEEEkeywords}}

\maketitle

\IEEEdisplaynontitleabstractindextext

%
\IEEEpeerreviewmaketitle

\ifCLASSOPTIONcaptionsoff
  \newpage
\fi

\renewcommand{\paragraph}[1]{\textbf{#1}}

	\section{Introduction}
	
	Models matter. Since the seminal work of AlexNet~\cite{krizhevsky2012imagenet} demonstrated the advantages of deep models over shallow ones on the ImageNet  challenge~\cite{russakovsky2015imagenet}, many novel deep convolutional neural network (CNN)~\cite{Lecun1989Backpropagation} models have been proposed and have significantly improved in performance, such as VGG~\cite{simonyan2014very}, Inception~\cite{szegedy2015going}, ResNet~\cite{He2016Deep}, and DenseNet~\cite{huang2016densely}. Fast, scalable, and end-to-end trainable CNNs have significantly advanced the field of computer vision in recent years, and particularly high-level vision problems. 
	
	Inspired by the successes of deep learning in high-level vision tasks, Dosovitskiy \etal~\cite{Dosovitskiy:2015Flownet} propose two CNN models for optical flow, \ie, FlowNetS and FlowNetC, and introduce a paradigm shift to this fundamental low/middle-level vision problem. Their work shows the feasibility of directly estimating optical flow from raw images using a generic U-Net CNN architecture~\cite{ronneberger2015u}. Although their performance is below the state of the art, FlowNetS and FlowNetC are the best among their contemporary real-time methods.

\begin{figure*}[t]
		\begin{center}
			\newcommand{\figwidth}{0.035\linewidth}
			\newcommand{\Figwidth}{0.48\linewidth}
			\newcommand{\Figheight}{0.375\linewidth}
			\newcommand{\shiftfigure}{\hspace{-3.5mm}}
			\begin{tabular}{cc}
			\shiftfigure
			\includegraphics[height = \Figheight]{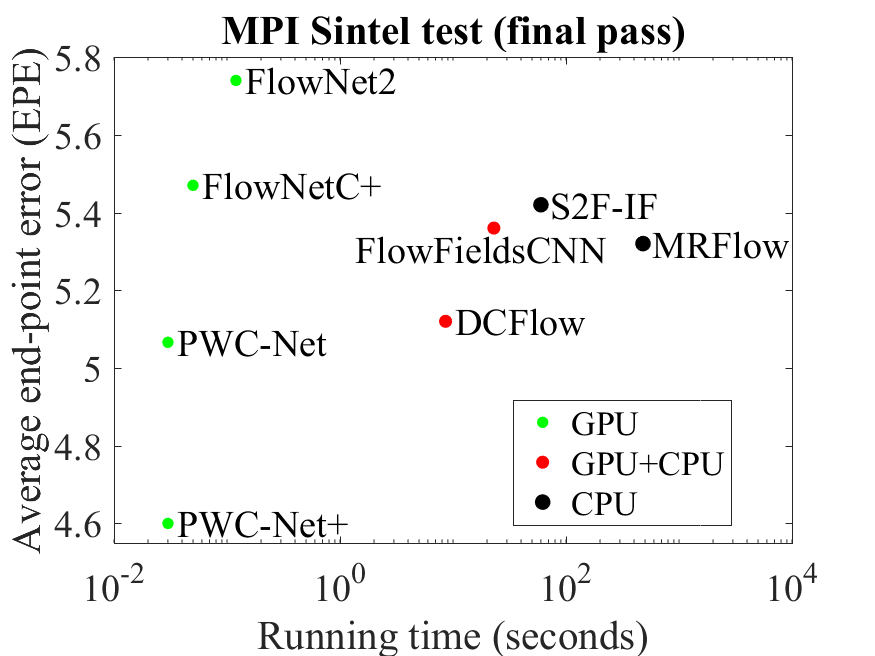} &
			\shiftfigure \shiftfigure
			\includegraphics[height = \Figheight]{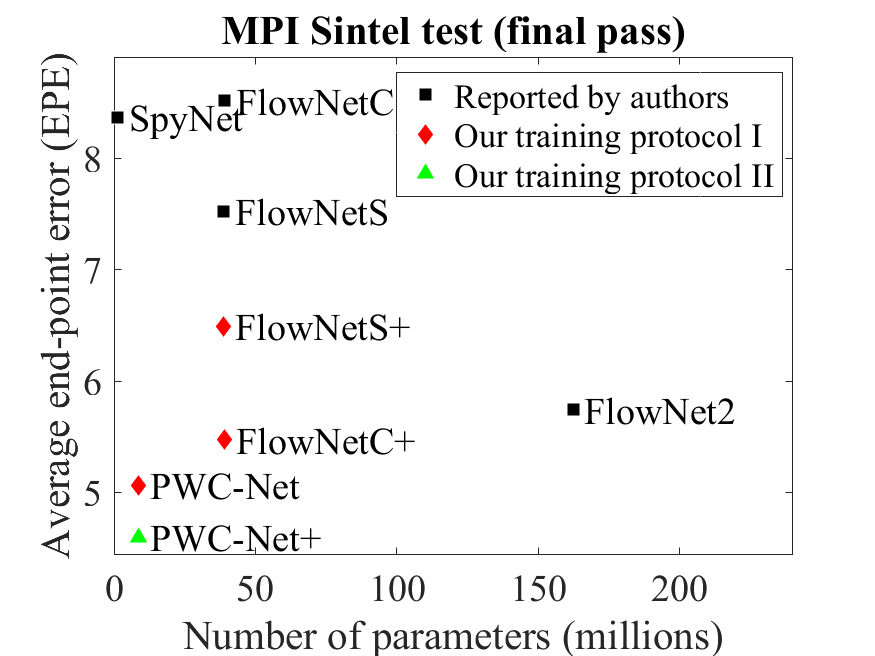}	
			\end{tabular}
			\caption{ 
				Left: PWC-Net outperforms all published methods on the MPI Sintel final pass benchmark in both accuracy and running time. 
				Right:  compared with previous end-to-end CNN models for flow, PWC-Net reaches the best balance between accuracy and size. The comparisons among PWC-Net, FlowNetS+, and FlowNetC+  show the improvements brought by the network architectures; all have been trained using the same training protocols.
				The comparisons,  FlowNetS vs. FlowNetS+, FlowNetC vs. FlowNetC+, and PWC-Net vs. PWC-Net+, show the improvements brought by training protocols. 
				\textbf{Both models and training matter.}
			}		
			\label{fig:memory:aepe}
		\end{center}
	\end{figure*}

	Recently, Ilg \etal~\cite{Ilg:2016:Flownet2} stacked one FlowNetC and several FlowNetS networks into a large model, called FlowNet2, which performs on par with state-of-the-art methods but runs much faster (Fig.~\ref{fig:memory:aepe}). 
	However, large models are more prone to the over-fitting problem, and as a result, the sub-networks of FlowNet2 have to be trained sequentially. Furthermore, FlowNet2 requires a memory footprint of 640MB and is not well-suited for mobile and embedded devices.
	
	SpyNet~\cite{Ranjan:2016:SpyNet} addresses the model size issue by combining deep learning with two classical optical flow estimation principles. SpyNet uses a spatial pyramid network and warps the second image toward the first one using the initial flow. The motion between the first and warped images is usually small. Thus SpyNet only needs a small network to estimate the motion from these two images. SpyNet performs on par with FlowNetC but below FlowNetS and FlowNet2. The reported results by FlowNet2 and SpyNet show a clear trade-off between accuracy and model size. 	
		
	\emph{Is it possible to both increase the accuracy and reduce the size  of a CNN model for optical flow estimation?} In principle, the trade-off between model size and accuracy imposes a fundamental limit for general machine learning algorithms. 
	However, we find that combining domain knowledge with deep learning can achieve both goals simultaneously for the particular problem of optical flow estimation.

	SpyNet shows the potential of combining classical principles with CNNs. However, we argue that its performance gap with FlowNetS and FlowNet2 is due to the partial use of classical principles. First, traditional optical flow methods often pre-process the raw images to extract {features} that are invariant to shadows or lighting changes~\cite{Baker:2011:DEO,weber1995robust}.
	Further, in the special case of stereo matching, a cost volume is a more discriminative representation of the disparity (1D flow) than raw images or features~\cite{Hosni2013Fast,Scharstein:2002:TEDS,Zbontar2016Stereo}. While constructing a full cost volume is computationally prohibitive for real-time optical flow estimation~\cite{Xu2017Accurate}, our work constructs a ``partial''
	cost volume by limiting the search range at each pyramid level. Linking different pyramid levels using a warping operation allows the estimation of large displacement flow.

	Our network, called PWC-Net, has been designed to make full use of these simple and well-established principles. It makes significant improvements in model size and accuracy over existing CNN models for optical flow (Fig.~\ref{fig:memory:aepe}). 
	PWC-Net is about 17 times smaller in size and 2 times faster in inferencing than FlowNet2. It is also easier to train than SpyNet and FlowNet2 and runs at about 35 frames per second (fps) on Sintel resolution ($1024\!\times\!436$) images. 
	It is the winning entry in the optical flow category of the first robust vision challenge. 

	However, it is imprecise or even misleading to conclude that the performance gains of PWC-Net come only from the new network architecture, because training matters as well.  If trained improperly, a good model may perform poorly. 
	CNNs were introduced in the late 80's~\cite{Lecun1989Backpropagation,Lecun1998Gradient}, but it took decades to figure out the details to train deep CNNs properly, such as dropout, ReLU units, batch normalization, and data augmentation~\cite{krizhevsky2012imagenet}. For optical flow, Dosovitskiy \etal \cite{Dosovitskiy:2015Flownet} report that FlowNetS outperforms FlowNetC. Ilg \etal \cite{Ilg:2016:Flownet2} show that using more iterations and dataset scheduling results in improved performance for both FlowNetS and FlowNetC. In particular, FlowNetC performs better than FlowNetS with the new training procedure. 
	For PWC-Net, we have used the same network architecture in the first version of our arXiv paper published in Sep.~2017, the second (CVPR) version in Nov.~2017, and the current one. The improvements over previous versions result solely from better training procedures. 
	
	In some sense, a straight forward comparison between PWC-Net and previous models is unfair, because the models have been trained differently. The potential of some models may be unfulfilled due to less than optimal training procedures. 
	To  fairly compare models, we retrain FlowNetC and FlowNetS, the sub-networks of FlwoNet2, using the same training protocol as PWC-Net. We observe significant performance improvements: the retrained FlowNetC is about 56\% more accurate on Sintel final than the previously trained one, although still 8\% less accurate than PWC-Net.
	A somewhat surprising result is that the retrained FlowNetC is about 5\% more accurate on Sintel final compared to the published FlowNet2 model, which has a much larger capacity. The last comparison clearly shows that better training procedures may be more effective at improving the performance of a basic model than increasing the model size, because larger models are usually harder to train\footnote{We cannot directly apply our training procedure to FlowNet2.}.
	The results show a complicated interplay between models and training, which requires careful experimental designs to identify the sources of performance gains.
	
	In this paper, we further improve the training procedures for PWC-Net. Specifically, adding KITTI and HD1K data during fine-tuning improves the average end-point error (EPE) of PWC-Net on Sintel final by 10\% to 4.60, which is better than all published methods. Fixing an I/O bug, which incorrectly loaded 22\% of the training data, leads to a~$\sim$20\% improvement on KITTI 2012 and 2015. At the time of writing, PWC-Net has the second lowest outlier percentage in non-occluded regions on KITTI 2015,  surpassed only by a recent scene flow method that uses stereo input and semantic information~\cite{Behl2017ICCV}. 
	
	To summarize, we make the following contributions:
	\begin{itemize}
\item We present a compact and effective CNN model for optical flow based on well-established principles. It performs robustly across four major flow benchmarks and is the winning entry in the optical flow category of the robust vision challenge. 
\item We compare FlowNetS, FlowNetC and PWC-Net trained using the same training procedure.  On Sintel final, the retrained FlowNetC is about 5\% more accurate than  the reported FlowNet2, which uses FlowNetC as a sub-network.
\item We further improve the training procedures for Sintel and fix an I/O bug for KITTI, both resulting in significant performance gain for the same PWC-Net network architecture. The newly trained model parameters and training protocols will be available on \url{https://github.com/NVlabs/PWC-Net}.
	\end{itemize}

\begin{figure*}[t]
	\begin{center}		
		\newcommand{\shiftfigure}{\hspace{0\linewidth}}
		\includegraphics[width=0.98\linewidth]{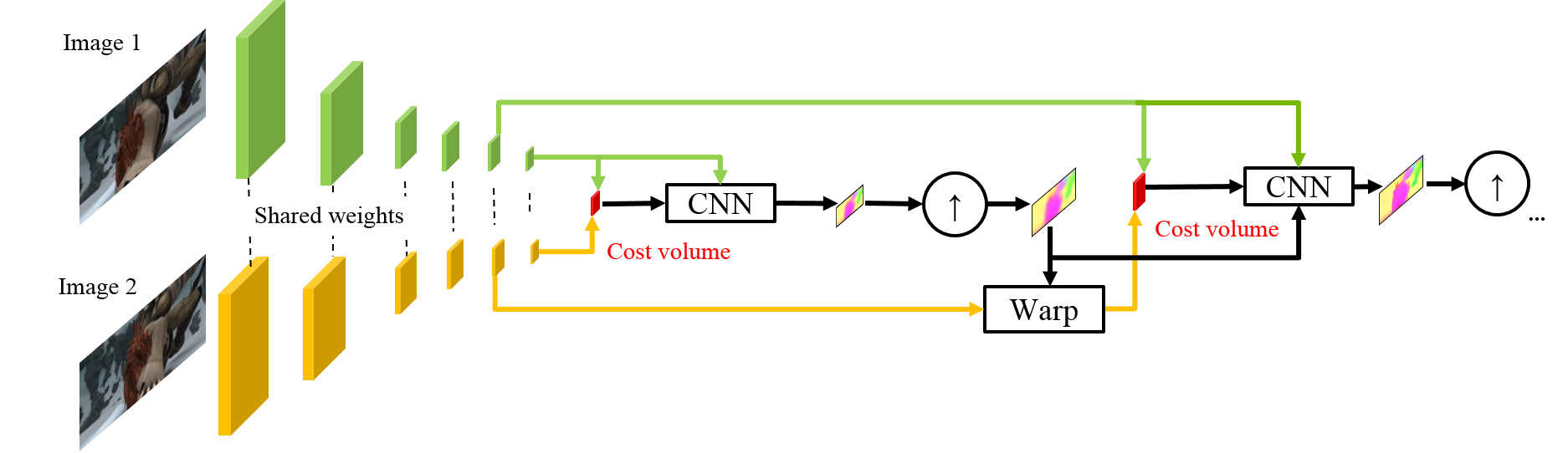} 
	\end{center}
	\caption{{\bf Network architecture of PWC-Net}. We only show the flow estimation modules at the top two levels. For the rest of the pyramidal levels, the flow estimation modules have the same structure as the second to top level.}
	\label{fig:network}
\end{figure*}

	\section{Previous Work}
	
	\paragraph{Variational approach.} 
	Horn and Schunck~\cite{Horn:1981:DO} pioneer the variational approach to optical flow by coupling the brightness constancy and spatial smoothness assumptions using an energy function. Black and Anandan~\cite{Black:1996:REMO} introduce a robust framework to deal with outliers, \ie, brightness inconstancy and spatial discontinuities. As it is computationally impractical to perform a full search, a coarse-to-fine, warping-based approach is often adopted~\cite{Bruhn:2005:CLG}. Brox \etal~\cite{Brox:2004:HAO} theoretically justify the warping-based estimation process.  The variational approach is the most popular framework for optical flow. However, it requires solving complex optimization problems and is computationally expensive for real-time applications. 
	
	Sun \etal~\cite{Sun:IJCV:2014} review the models, optimization, and implementation details for methods derived from Horn and Schunck. One surprising finding is that the original Horn and Schunck objective, when optimized using modern techniques and implementation details, performs competitively against contemporary state of the art. Thus, it is critical to separate the contributions from the objective and the optimization. Our work shows that it is also critical to separate the contributions from the CNN models and the training procedures.
	
	One conundrum for the coarse-to-fine approach is small and fast moving objects that disappear at coarse levels. To address this issue, Brox and Malik~\cite{Brox:LDOF:2011} embed feature matching into the variational framework, which is further improved by follow-up methods~\cite{Weinzaepfel:2013:DeepFlow,Xu:2012:MDP}. In particular, the EpicFlow method~\cite{EpicFlow} can effectively interpolate sparse matches to dense optical flow and is widely used as a post-processing method~\cite{Bai2016Exploiting,Bailer_2017_CVPR,Chen2016Full,Hu2016Efficient,Xu2017Accurate,yang2017s2f}. Zweig and Wolf~\cite{Zweig_2017_CVPR} use CNNs for sparse-to-dense interpolation and obtain consistent improvement over EpicFlow.
	
	Most top-performing methods use CNNs as a component in their system. For example, DCFlow~\cite{Xu2017Accurate},  the best published method on MPI Sintel final pass so far, learns CNN features to construct a full cost volume and uses sophisticated post-processing techniques, including EpicFlow, to estimate the optical flow. 
	The next-best method, FlowFieldsCNN~\cite{Bailer_2017_CVPR}, learns CNN features for sparse matching and densifies the matches by EpicFlow. 
	The third-best method, MRFlow~\cite{Wulff2017Optical} uses a CNN to classify a scene into rigid and non-rigid regions and estimates the geometry and camera motion for rigid regions using a plane + parallax formulation. However, none of them are real-time or end-to-end trainable.

	\beforePara
	\paragraph{Early work on learning optical flow.}
	There is a long history of learning optical flow before the deep learning era.
	Simoncelli and Adelson~\cite{Simoncelli:1991:Probability} study the data matching errors for optical flow.
	Freeman \etal~\cite{Freeman:2000:LLLV} learn parameters of an MRF model for image motion using synthetic blob world examples.
	Roth and Black~\cite{Roth:2007:SSO} study the spatial statistics of optical flow using sequences generated from depth maps. 
	Sun \etal~\cite{Sun:2008:LOF} learn a full model for optical flow, but the learning has been limited to a few training sequences~\cite{Baker:2011:DEO}. 
	Li and Huttenlocker~\cite{Li2008Learning} use stochastic optimization to tune the parameters for the Black and Anandan method~\cite{Black:1996:REMO}, but the number of parameters learned is limited. Wulff and Black~\cite{Wulff:2015:PCA} learn PCA motion basis of optical flow estimated by GPUFlow~\cite{Werlberger:2009:AHOF} on real movies. Their method is fast but produces over-smoothed flow.
	
	\beforePara
	\paragraph{Recent work on learning optical flow.}
	Inspired by the success of CNNs on high-level vision tasks~\cite{krizhevsky2012imagenet}, Dosovitskiy \etal~\cite{Dosovitskiy:2015Flownet}
	construct two CNN networks, FlowNetS and FlowNetC, for estimating optical flow based on the U-Net denoising autoencoder~\cite{ronneberger2015u}.
	The networks are pre-trained on a large synthetic FlyingChairs dataset but  can surprisingly capture the motion of fast moving objects on the Sintel dataset.
	The raw output of the network, however, contains large errors in smooth background regions and requires variational refinement~\cite{Brox:LDOF:2011}. 
	Mayer \etal~\cite{Mayer:2016:Large}  apply the FlowNet architecture to disparity and scene flow estimation.
	Ilg \etal~\cite{Ilg:2016:Flownet2} stack several basic FlowNet models into a large one, \ie, FlowNet2, which performs on par with state of the art on the Sintel benchmark. 
	Ranjan and Black~\cite{Ranjan:2016:SpyNet} develop a compact spatial pyramid network, called SpyNet. SpyNet achieves similar performance as the FlowNetC model on the Sintel benchmark, which is good but not state-of-the-art.
	
	Another interesting line of research takes the unsupervised learning approach.  Memisevic and Hinton~\cite{Memisevic:2007:Unsupervised} propose the gated restricted Boltzmann machine to learn image transformations in an unsupervised way. Long \etal~\cite{Long:2016:Learning} learn CNN models for optical flow by interpolating frames. Yu \etal~\cite{YuHD16} train models to minimize a loss term that combines a data constancy term with a spatial smoothness term. While inferior to supervised approaches on datasets with labeled training data, existing unsupervised methods can be used to (pre-)train CNN models on unlabeled data~\cite{Lai2017SemiFlowGAN}.  
	
	\beforePara
	\paragraph{Cost volume.} A cost volume stores the data matching costs for associating a pixel with its corresponding pixels at the next frame~\cite{Hosni2013Fast}. Its computation and processing are standard components for stereo matching, a special case of optical flow. Recent methods~\cite{Chen2016Full,Dosovitskiy:2015Flownet,Xu2017Accurate} investigate cost volume processing for optical flow. All build the full cost volume at a single scale, which is both computationally expensive and memory intensive. By contrast, our work shows that constructing a partial cost volume at multiple pyramid levels leads to both effective and efficient models.

\begin{figure*}[t]
	\begin{center}		
		\newcommand{\shiftfigure}{\hspace{0\linewidth}}
		\includegraphics[width=0.98\linewidth]{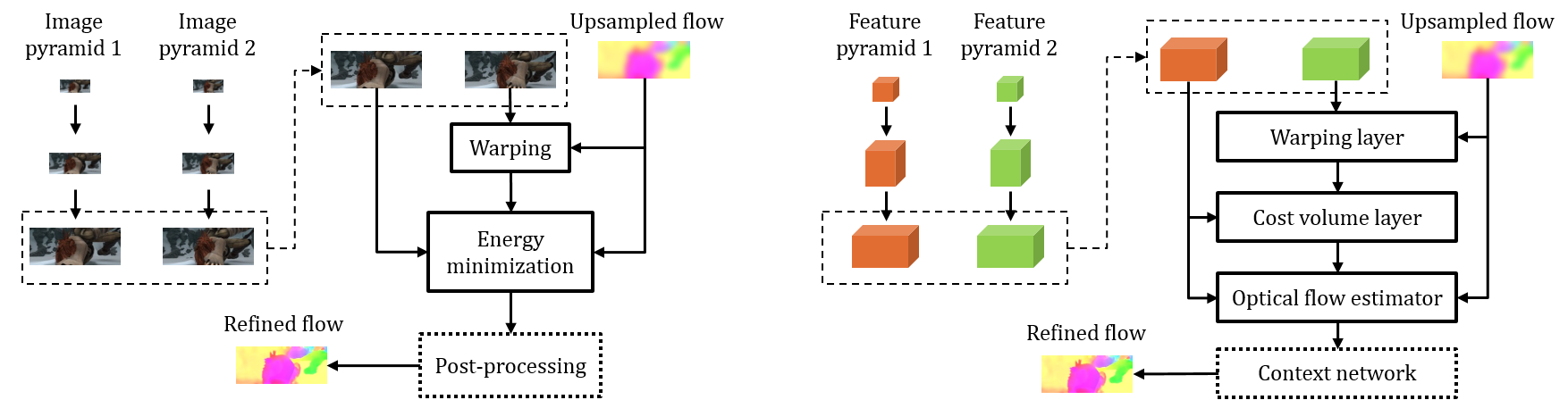} 
	\end{center}
	\caption{{\bf Traditional coarse-to-fine approach vs. PWC-Net.} Left: Image pyramid and refinement at one pyramid level by the energy minimization approach~\cite{Black:1996:REMO,Brox:2004:HAO,Horn:1981:DO,Sun:IJCV:2014}.  Right: Feature pyramid and refinement at one pyramid level by PWC-Net. PWC-Net  warps features of the second image using the upsampled flow,  computes a cost volume, and process the cost volume using CNNs. Both post-processing and context network are optional in each system. The arrows indicate the direction of flow estimation and pyramids are constructed in the opposite direction.	Please refer to the text for details about the network. 
	}
	\label{fig:c2f:pwc}
\end{figure*}
	
	\beforePara
	\paragraph{Datasets.} 
	Unlike many other vision tasks, it is extremely difficult to obtain ground truth optical flow on real-world sequences. Early work on optical flow mainly relies on synthetic datasets~\cite{Barron:1994:PO}, \eg, the famous ``Yosemite''. Methods may over-fit to the synthetic data and do not perform well on real data~\cite{Liu:2008:HAMA}. Baker \etal~\cite{Baker:2011:DEO} capture real sequences under both ambient and UV lights in a controlled lab environment to obtain ground truth, but the approach does not work for outdoor scenes. Liu \etal~\cite{Liu:2008:HAMA} use human annotations to obtain ground truth motion for natural video sequences, but the labeling process is time-consuming. 
	
	KITTI and Sintel are currently the most challenging and widely-used benchmarks for optical flow. The KITTI benchmark is targeted for autonomous driving applications and its semi-dense ground truth is collected using LIDAR~\cite{Geiger:2012:KITTI}. The 2012 set only consists of static scenes.  The 2015 set is extended to dynamic scenes via human annotations and more challenging to existing methods because of the large motion, severe illumination changes, and occlusions~\cite{Menze2015CVPR}. 
	The Sintel benchmark~\cite{Butler:ECCV:2012} is created using the open source graphics movie ``Sintel'' with two passes, clean and final. The final pass contains strong atmospheric effects, motion blur,  and camera noise, which cause severe problems to existing methods.  
	All published, top-performing methods~\cite{Bailer_2017_CVPR,Wulff2017Optical,Xu2017Accurate}  rely heavily on traditional techniques. 
	PWC-Net is the first fully end-to-end CNN model that outperforms all published methods on both the KITTI 2015 and Sintel final pass benchmarks. 

	\beforePara
	\paragraph{CNN models for dense prediction tasks in  vision.}
	The denoising autoencoder~\cite{vincent2008extracting} has been commonly used for dense prediction tasks in computer vision, especially with skip connections~\cite{ronneberger2015u} between the encoder and decoder. Recent work~\cite{chen2017deeplab,yu2015multi} shows that dilated convolution layers can better exploit contextual information and refine details for semantic segmentation. Here we use dilated convolutions to integrate contextual information for optical flow and obtain moderate performance improvement. The DenseNet architecture~\cite{huang2016densely,Jegou:2016:Densenet} directly connects each layer to every other layer in a feedforward fashion and has been shown to be more accurate and easier to train than traditional CNN layers in image classification tasks. We test this idea for dense optical flow prediction.

	\section{Approach}
	
	We start from an overview of the network architecture of PWC-Net, as shown in Figure~\ref{fig:network}. PWC-Net first builds a feature pyramid from the two input images. At the top level of the pyramid, PWC-Net constructs a cost volume by comparing features of a pixel in the first image with corresponding features in the second image. As the top level is of small spatial resolution, we can construct the cost volume using a small search range. The cost volume and features of the first image are then fed to a CNN to estimate the flow at the top level. PWC-Net then upsamples and rescales the estimated flow to the next level. At the second to top level, PWC-Net warps features of the second image toward the first using the upsampled flow, and then constructs a cost volume using features of the first image and the warped features. As warping compensates the large motion, we can still use a small search range to construct the cost volume. The cost volume, features of the first image, and the upsampeld flow are fed to a CNN to estimate flow at the current level, which is then upsampled to the next (third) level. The process repeats until the desired level. 
	
	As PWC-Net has been designed using classical principles from optical flow, it is informative to compare the key components of PWC-Net with the traditional coarse-to-fine approaches~\cite{Black:1996:REMO,Brox:2004:HAO,Horn:1981:DO,Sun:IJCV:2014} in Figure~\ref{fig:c2f:pwc}. First, as raw images are variant to shadows and lighting changes~\cite{Brox:2004:HAO,Sun:IJCV:2014},  we replace the fixed image pyramid with learnable feature pyramids. Second, we take the warping operation from the traditional approach as a layer in our network to estimate large  motion. Third, as the cost volume is a more discriminative representation of the optical flow than raw images, our network has a layer to construct the cost volume, which is then processed by CNN layers to estimate the flow. The warping and cost volume layers have no learnable parameters and  reduce the model size. Finally, a common practice by the traditional methods is to post-process the optical flow  using contextual information, such as median filtering~\cite{Wedel:2008:ITVL1} and bilateral filtering~\cite{Xiao:2006:BFO}. Thus, PWC-Net uses a context network to exploit contextual information to refine the optical flow. Compared with energy minimization, the CNN models are computationally more efficient. 
	
	Next, we will explain the main ideas for each component, including pyramid feature extractor, optical flow estimator, and context networks. Please refer to the appendix for details of the networks.

	\beforePara
	\paragraph{Feature pyramid extractor.}	Given two input images $\mv{I}_1$ and $\mv{I}_2$, we generate $L$-level pyramids of feature representations, with the bottom (zeroth) level being the input images, \ie,  $\mv{c}^0_{t} = \mv{I}_t$.  
	To generate feature representation at the $l$th layer, $\mv{c}^l_t$, we use layers of convolutional filters to downsample the features at the $l\!-\!1$th pyramid level,  $\mv{c}^{l\!-\!1}_t$, by a factor of $2$.
	From the first to the sixth levels, the number of feature channels are respectively $16$, $32$, $64$, $96$, $128$, and $192$. 
	
	\beforePara
	\paragraph{Warping layer.} At the $l$th level, we first upsample by a factor of 2 and rescale the estimated flow from the $l\!+\!1$th level, $\mv{w}^{l\!+\!1}$, to the current level. We then  warp features of the second image toward the first image using the upsampled flow: 
	\beforeEqn
	\begin{align}
	\mv{c}^{l}_w (\mv{x}) = \mv{c}^{l}_2 \big (\mv{x} + 2\times\mathrm{up}_2 (\!\mv{w}^{l\!+\!1}) (\mv{x}) \big ),
	\end{align}
	where $\mv{x}$ is the pixel index and $\mathrm{up}_2$ denote the $\times2$ upsampling operator. We use bilinear interpolation to implement the warping operation and compute the gradients to the input CNN features and flow for backpropagation according to~\cite{Ilg:2016:Flownet2,Jaderberg:2015:Spatial}. For non-translational motion, warping can compensate for some geometric distortions and put image patches at the right scale.  Note that there is no upsampled flow at the top pyramid level and the warped features are the same as features of the second image, \ie, 	$\mv{c}^{L}_w \!=\! \mv{c}^{L}_2$. 
	
	\beforePara
	\paragraph{Cost volume layer.} 
	Next, we use the features to construct a cost volume that stores the matching costs for associating a pixel with its corresponding pixels at the next frame~\cite{Hosni2013Fast}. We define the matching cost as the correlation~\cite{Dosovitskiy:2015Flownet,Xu2017Accurate} between features of the first image and warped features of the second image: 
	\beforeEqn
	\begin{align}
	\!\!\!\mv{cv}^l(\mv{x}_1, \mv{x}_2) \!=\! \frac{1}{N} \left ( \mv{c}^l_1(\mv{x}_1) \right )^\mathsf{T} \mv{c}^l_w(\mv{x}_2),\!\!\!\!\!
	\end{align} 
	where $\mathsf{T}$ is the transpose operator and $N$ is the length of the column vector $\mv{c}^l_1(\mv{x}_1)$. 
	For an $L$-level pyramid setting, we only need to compute a partial cost volume with a limited range of $d$ pixels, \ie, $|\mv{x}_1 -\mv{x}_2|_{\infty}\!\leq\!d$. 
	A one-pixel motion at the top level corresponds to $2^{L\!-\!1}$ pixels at the full resolution images. Thus we can set $d$ to be small.  The dimension of the 3D cost volume is $d^2 \!\times\! H^l \!\times\! W^l$, where $H^l $ and $W^l $ denote the height and width of the $l$th pyramid level, respectively.

	\beforePara
	\paragraph{Optical flow estimator.} 
	It is a multi-layer CNN. Its input are the cost volume, features of the first image, and upsampled optical flow and its output is the flow $\mv{w}^{l}$ at the $l$th level. The numbers of feature channels at each convolutional layers are respectively 128, 128, 96, 64, and 32, which are kept fixed at all pyramid levels. The estimators at different levels have their own parameters instead of sharing the same parameters. This estimation process is repeated until the desired level, $l_0$.

	The estimator architecture can be enhanced with DenseNet connections~\cite{huang2016densely}. The inputs to every convolutional layer are the output of and the input to its previous layer. DenseNet has more direct connections than traditional layers and leads to significant improvement in image classification. We test this idea for dense flow prediction.

	\beforePara
	\paragraph{Context network.}	
	Traditional flow methods often use contextual information to post-process the flow. Thus we employ a sub-network, called the context network, to effectively enlarge the receptive field size of each output unit at the desired pyramid level.  It takes the estimated flow and features of the second last layer from the optical flow estimator and outputs a refined flow, $\hat{\mv{w}}^{l_0}_{\Theta}(\mv{x})$.

	The context network is a feed-forward CNN and its	design is based on dilated convolutions~\cite{yu2015multi}. It consists of 7 convolutional layers. The spatial kernel for each convolutional layer is 3$\times$3. These layers have different dilation constants. A convolutional layer with a dilation constant $k$ means that an input unit to a filter in the layer are $k$-unit apart from the other input units to the filter in the layer, both in vertical and horizontal directions. Convolutional layers with large dilation constants enlarge the receptive field of each output unit without incurring a large computational burden. From bottom to top, the dilation constants are $1, 2, 4, 8, 16, 1$, and $1$.

	\paragraph{Training loss.}
	Let $\Theta$ be the set of all the learnable parameters in our final network, which includes the feature pyramid extractor and the optical flow estimators at different pyramid levels (the warping and cost volume layers have no learnable parameters).
	Let $\mv{w}^l_{\Theta}$ denote the flow field at the $l$th pyramid level predicted by the network, and $\mv{w}^l_{\textrm{GT}}$ the corresponding supervision signal. 
	We use the same multi-scale training loss proposed in FlowNet~\cite{Dosovitskiy:2015Flownet}:
	\beforeEqn
	\begin{align} 
	\mathcal{L}(\Theta)  \!=\! \sum^L_{l=l_0} \alpha_l \sum_{\mv{x}} |\mv{w}^l_{\Theta}(\mv{x})\!-\! \mv{w}^l_{\textrm{GT}}(\mv{x})|_2 \!+\! \gamma |\Theta|^2_2,
	\label{eq:multi:loss}
	\end{align}
	where $|\cdot|_2$ computes the L2 norm of a vector and the second term regularizes parameters of the model. Note that if the context network is used at the $l_0$th level, $\mv{w}^{l_0}_{\Theta}$ will be replaced by the output of the context network, $\hat{\mv{w}}^{l_0}_{\Theta}(\mv{x})$.
	For fine-tuning, we use the following robust training loss: 
	\beforeEqn
	\begin{align}
	\mathcal{L}(\Theta)  \!=\! \sum^L_{l=l_0} \alpha_l\! \sum_{\mv{x}}  \left ( |\mv{w}^l_{\Theta}(\mv{x})\!-\! \mv{w}^l_{\textrm{GT}}(\mv{x})| \!+\!\epsilon \right) ^q \!+\! \gamma |\Theta|^2_2
	\label{eq:multi:loss:lq}
	\end{align}
	where $|\cdot|$ denotes the L1 norm, $q\!<\!1$ gives less penalty to outliers, and $\epsilon$ is a small constant.

	\section{Experimental Results}
	
	\paragraph{Implementation details.}
	The weights in the training loss~\eqref{eq:multi:loss} are set to be $\alpha_6\!=\!0.32, \alpha_5\!=\!0.08, \alpha_4\!=\!0.02, \alpha_3\!=\!0.01$, and $\alpha_2\!=\!0.005$. 
	The trade-off weight $\gamma$ is set to be $0.0004$.
	We scale the ground truth flow by $20$ and downsample it to obtain the supervision signals at different levels. Note that we do not further scale the supervision signal at each level, the same as~\cite{Dosovitskiy:2015Flownet}.  As a result, we need to scale the upsampled flow at each pyramid level for the warping layer. For example, at the second level, we scale the upsampled flow from the third level by a factor of $5$ ($\!=\!20/4$) before warping features of the second image.  
	We use a 7-level pyramid ($L=6$), consisting of 6 levels of CNN features and the input images as the bottom level. We set the desired level $l_0$ to be 2, \ie, our model outputs a quarter resolution optical flow and uses bilinear interpolation to obtain the full-resolution optical flow.  	We use a search range of $4$ pixels to compute the cost volume at each level.

	We first train the models using the FlyingChairs dataset in Caffe~\cite{jia2014caffe} using the $S_{long}$ learning rate schedule introduced in~\cite{Ilg:2016:Flownet2}, \ie, starting from $0.0001$ and reducing the learning rate by half at $0.4$M, $0.6$M, $0.8$M, and $1$M iterations. The data augmentation scheme is the same as that in~\cite{Ilg:2016:Flownet2}. We crop $448\times384$ patches during data augmentation and use a batch size of 8. We then fine-tune the models on the FlyingThings3D dataset using the $S_{fine}$ schedule~\cite{Ilg:2016:Flownet2} while excluding image pairs with extreme motion (magnitude larger than 1000 pixels).  The cropped image size is  $768\times384$ and the batch size is 4. Finally, we fine-tune the models using the Sintel and KITTI training sets and will explain the details below.

	\begin{figure}[h]
		\begin{center}	
			\includegraphics[width=0.95\linewidth]{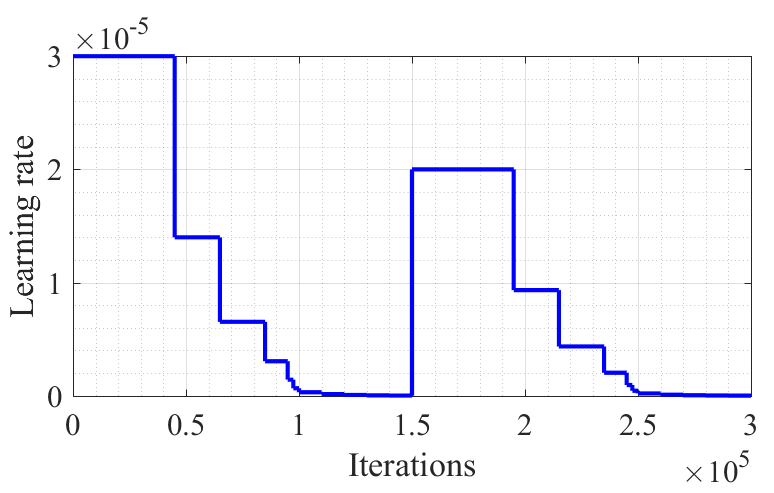} 
			\includegraphics[width=0.95\linewidth]{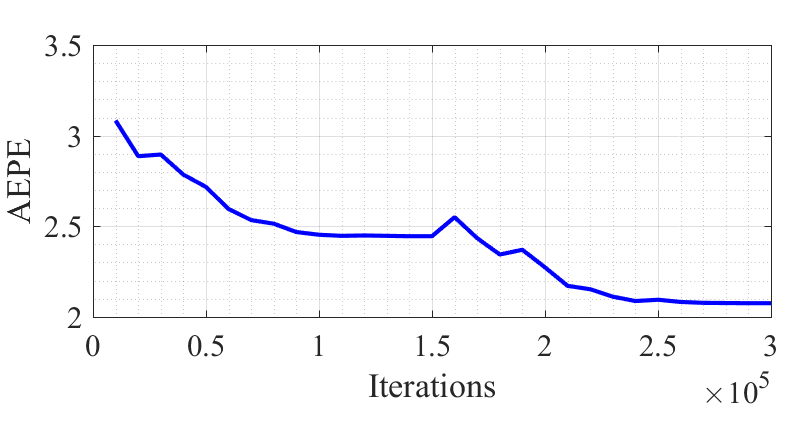} 	
		\end{center}
		\caption{Top: learning rate schedule for fine-tuning (the step values for the first $10^5$ iterations were provided by Eddy Ilg). 
			Bottom: average end-point error (EPE) on the final pass of the Sintel training set. 
			We disrupt the learning rate for a better minimum, which has better accuracy in both the training and the test sets.}
		\label{fig:ft:lr}
	\end{figure}

	\subsection{Main Results}
	
	\subsubsection{MPI Sintel.}
	When fine-tuning on Sintel, we crop $768\times384$ image patches, add horizontal flip, and do not add additive Gaussian noise during data augmentation. The batch size is 4. 
	We use the robust loss function in Eq.~\eqref{eq:multi:loss:lq} with $\epsilon=0.01$ and $q=0.4$. 
	We disrupt the learning rate, as shown in Fig.~\ref{fig:ft:lr}, which empirically improves both the training and test performance.
	We test two schemes of fine-tuning. The first one, PWC-Net-ft, uses the clean and final passes of the Sintel training data throughout the fine-tuning process. The second one, PWC-Net-ft-final, uses only the final pass for the second half of fine-tuning. We test the second scheme because DCFlow learns the features using only the final pass of the training data. Thus we test the performance of PWC-Net when the final pass of the training data is given more weight. We refer to the latter scheme as our \textbf{training protocol I}, which we will use later to train other models. 

	\begin{figure*}[th]
		\footnotesize
		\begin{center}
			\newcommand{\figwidth}{0.18\linewidth}
			\newcommand{\Figwidth}{0.32\linewidth}
			\newcommand{\shiftfigure}{\hspace{0mm}}
			\begin{tabular}{cccc}
				\shiftfigure \hspace{-1em}\subfigimg[width = \Figwidth]{Frame 46 of ``Market\_5'' (training, clean)}{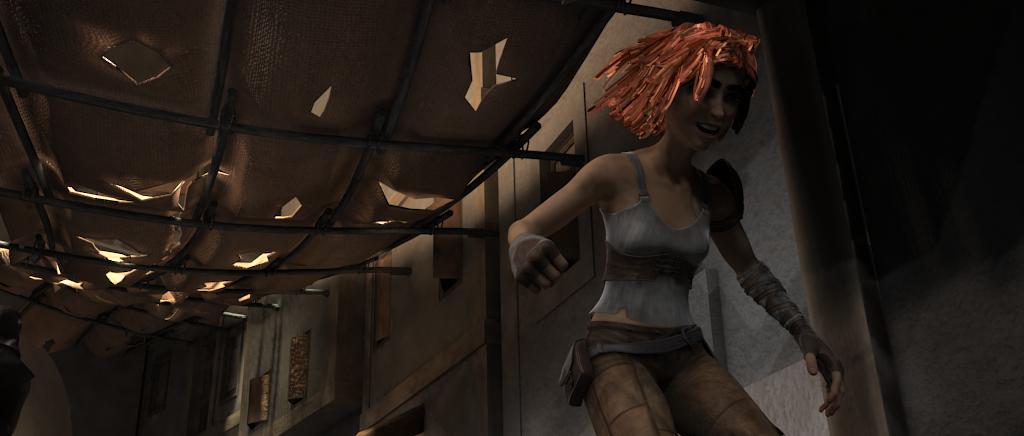} &
				\shiftfigure \hspace{-1em}\subfigimg[width = \Figwidth]{Ground truth}{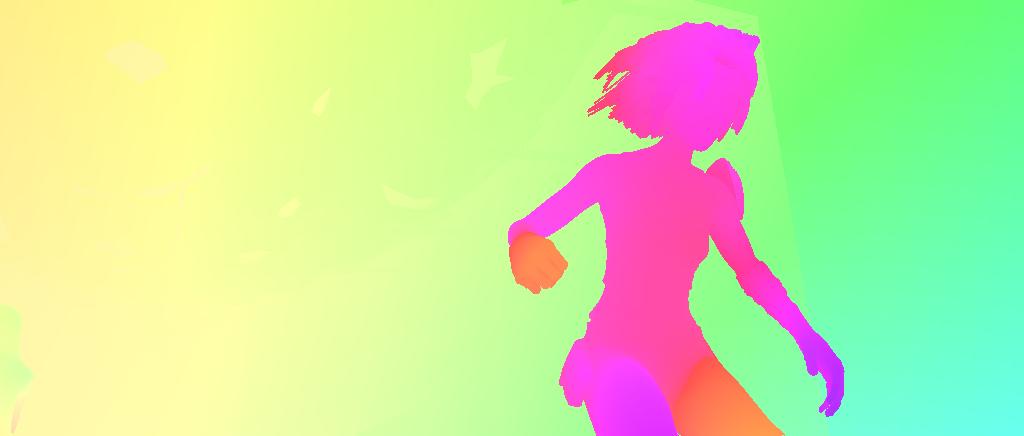} &
				\shiftfigure \hspace{-1em}\subfigimg[width = \Figwidth]{W/o context}{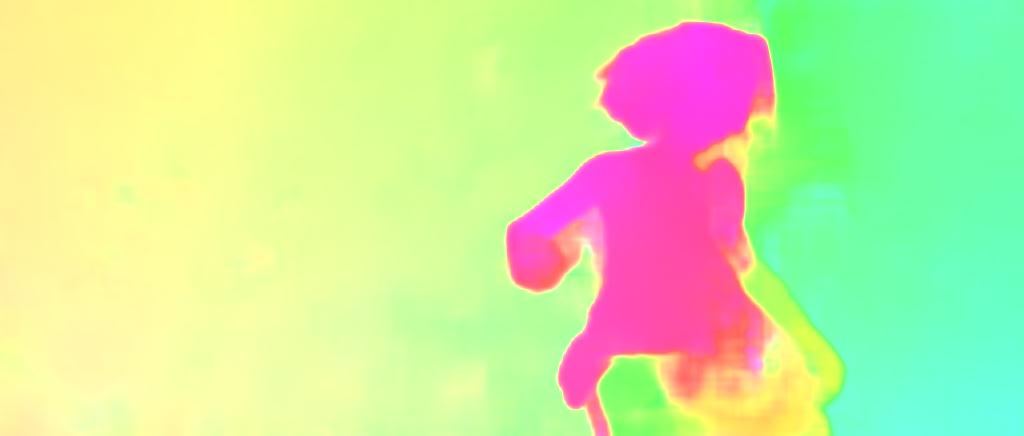} \\
				\shiftfigure \hspace{-1em}\subfigimg[width = \Figwidth]{W/o DenseNet }{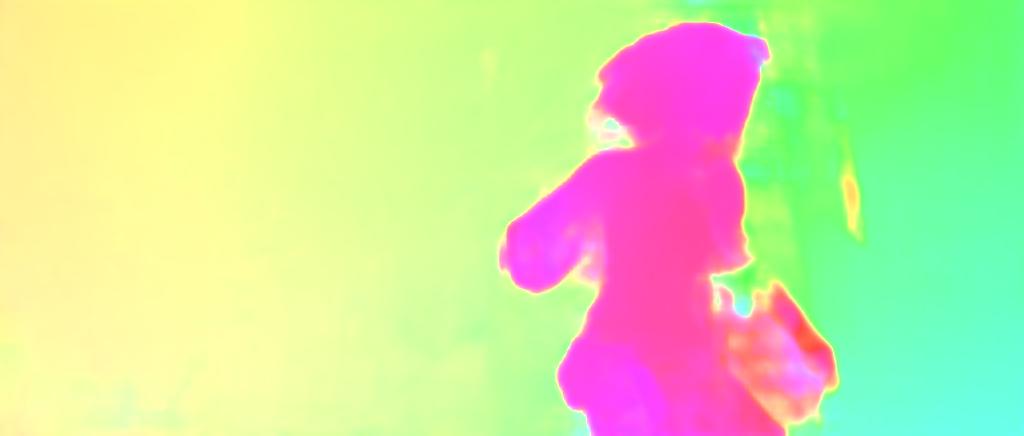} &
				\shiftfigure \hspace{-1em}\subfigimg[width = \Figwidth]{PWC-Net}{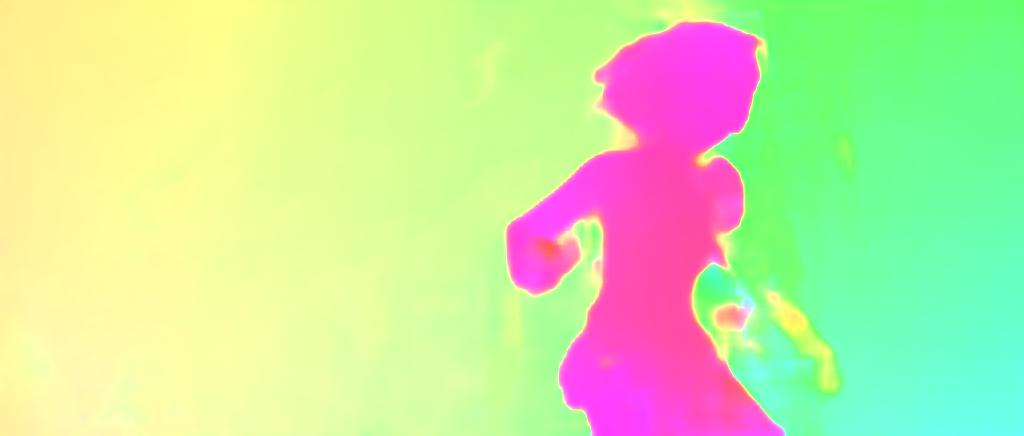} &
				\shiftfigure \hspace{-1em}\subfigimg[width = \Figwidth]{PWC-Net-Sintel-ft}{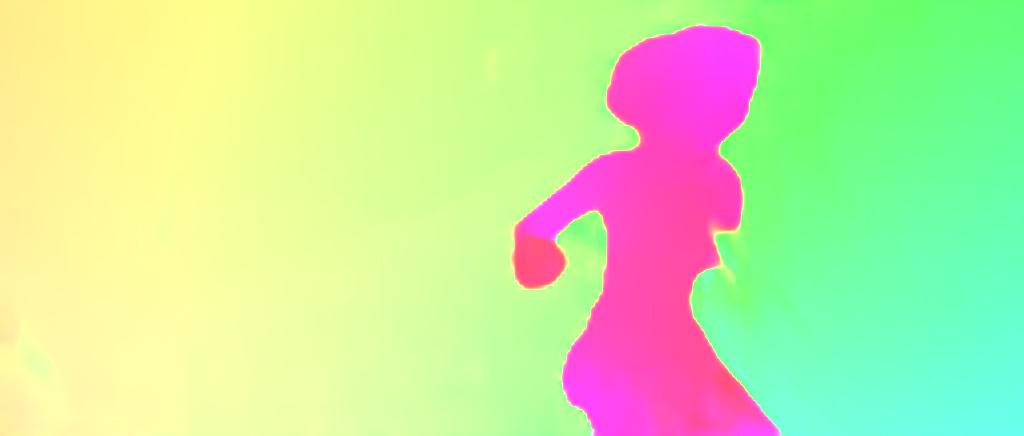}  \\	
				\shiftfigure \hspace{-1em}\subfigimg[width = \Figwidth]{Frame 5 of  ``Ambush\_3'' (test, final)}{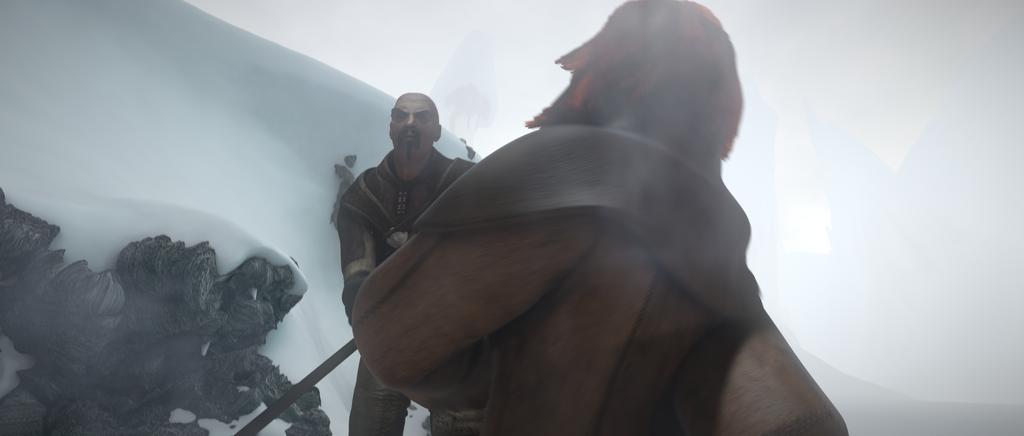} &
				\shiftfigure \hspace{-1em}\subfigimg[width = \Figwidth]{Frame 6}{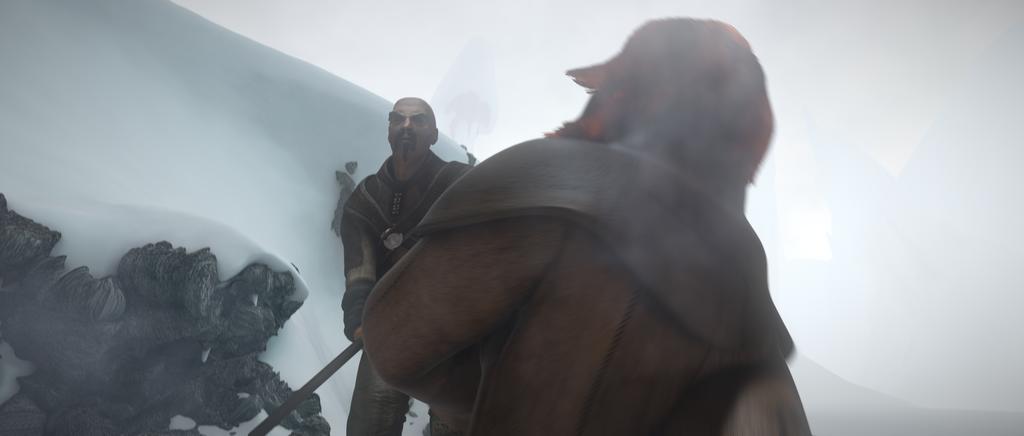} &
				\shiftfigure \hspace{-1em}\subfigimg[width = \Figwidth]{W/o context}{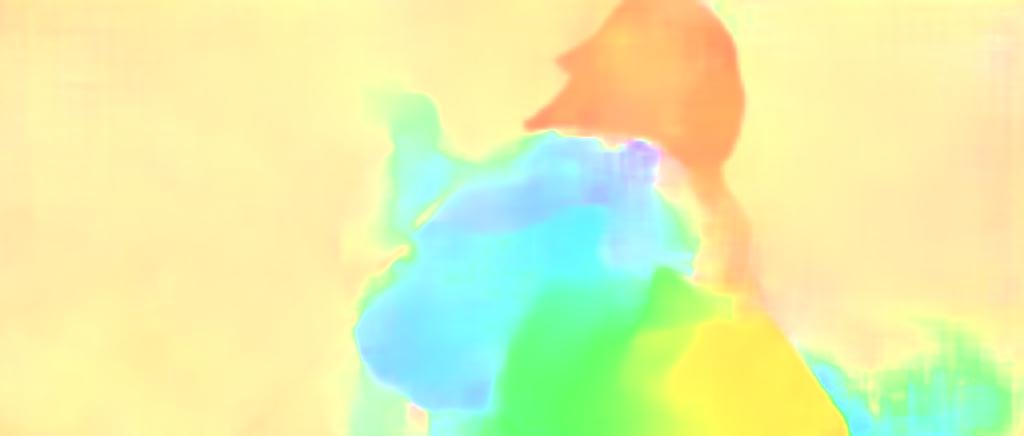} \\
				\shiftfigure \hspace{-1em}\subfigimg[width = \Figwidth]{W/o DenseNet }{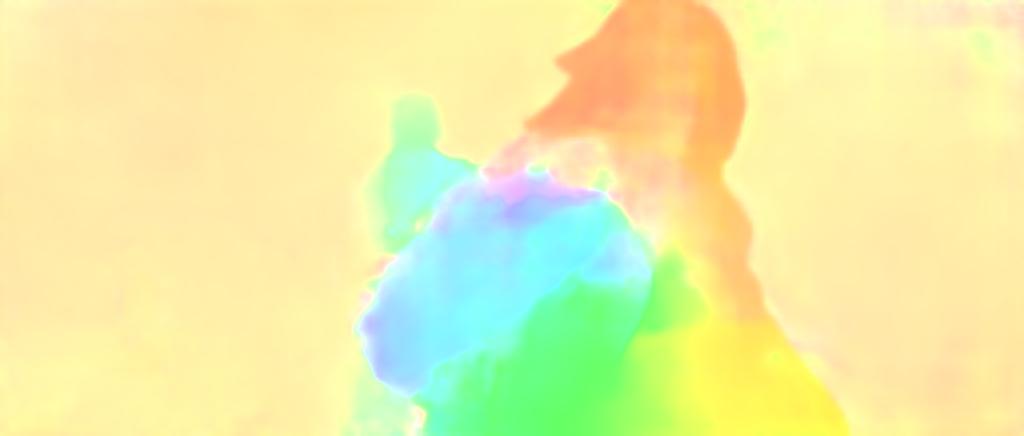} &
				\shiftfigure \hspace{-1em}\subfigimg[width = \Figwidth]{PWC-Net}{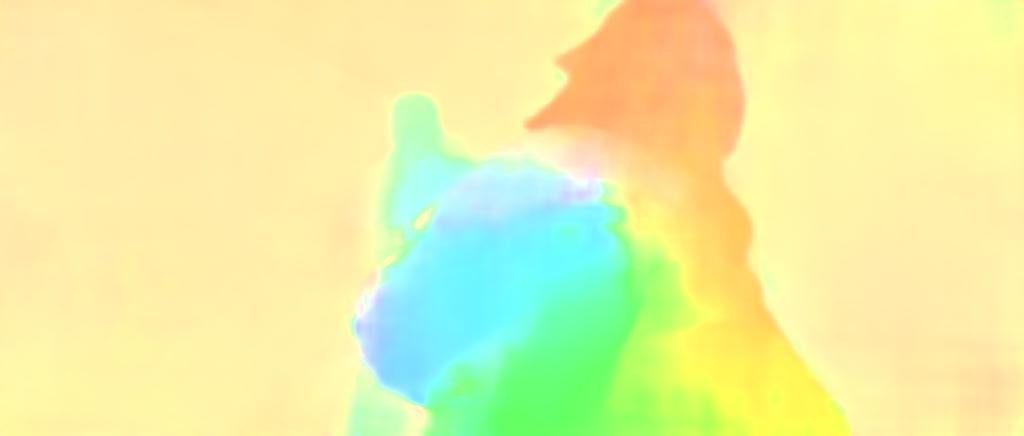} &
				\shiftfigure \hspace{-1em}\subfigimg[width = \Figwidth]{PWC-Net-Sintel-ft}{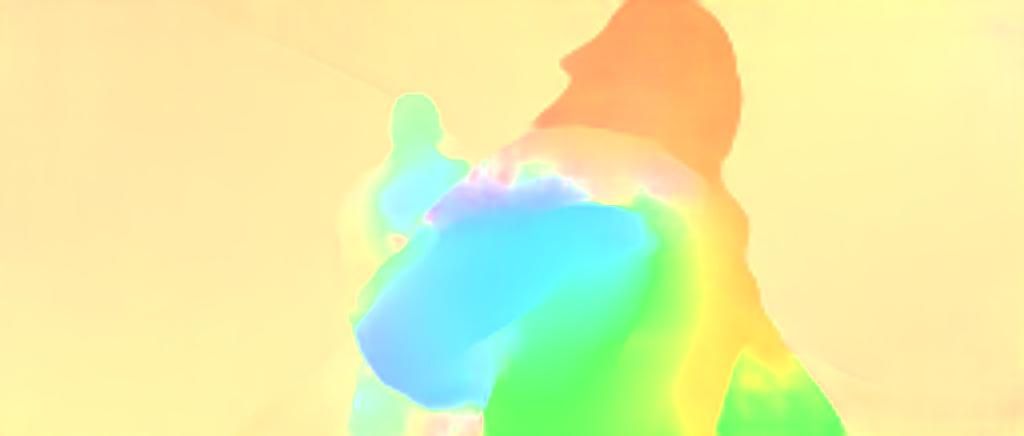}  \\		
			\end{tabular}		
		\end{center}
		\vspace{-4mm}
		\caption{Results on Sintel \emph{training} and \emph{test} sets. Context network, DenseNet connections, and fine-tuning all improve the results. Small and rapidly moving objects, \eg, the left arm in ``Market\_5'', are still challenging to the pyramid-based PWC-Net.
		}
		\label{fig:sintel:models}
	\end{figure*}
	
	PWC-Net has lower average end-point error (EPE) than many recent methods on the final pass of the MPI-Sintel benchmark (Table~\ref{tab:result:sintel}). Further, PWC-Net is the fastest among all the top-performing methods (Fig.~\ref{fig:memory:aepe}). We can further reduce the running time by dropping the DenseNet connections. The resulting PWC-Net-small model is about 5\% less accurate but 40\% faster than PWC-Net.

	PWC-Net is less accurate than traditional approaches on the clean pass. Traditional methods often use  image edges to refine  motion boundaries, because the two  are perfectly aligned in the clean pass. However, image edges in the final pass are  corrupted by motion blur, atmospheric changes, and noise. Thus, the final pass is more realistic and challenging. The results on the final and clean sets suggest that PWC-Net may be better suited for real images, where the image edges are often corrupted.

	PWC-Net has higher errors on the training set but lower errors on the test set than FlowNet2, suggesting that PWC-Net may have a more appropriate capacity for this task. Table~\ref{tab:result:sintel:comparison} summarizes  errors in different regions. PWC-Net performs relatively better in regions with large motion and away from the motion boundaries, probably because it has been trained using only data with large motion. Figure~\ref{fig:sintel:models} shows the visual results of different variants of PWC-Net on the training and test sets of MPI Sintel. PWC-Net can recover sharp motion boundaries but may fail on small and rapidly moving objects, such as the left arm in ``Market\_5''.

	\begin{table}[ht]
		\caption{Average EPE results on MPI Sintel set.  ``-ft'' means fine-tuning on the MPI Sintel \emph{training} set and the numbers in the parenthesis are results on the data the methods have been fine-tuned on. ft-final gives more weight to the final pass during fine-tuning. FlowNetC2 has been trained using the same procedure as PWC-Net-ft-final.} 	\vspace{-1mm}
		\label{tab:result:sintel}
		\footnotesize
		\centering
		\begin{tabular}{lcccccc} \\
			\multirow{2}*{Methods} & \multicolumn{2}{c}{Training}  & \multicolumn{2}{c}{Test} & Time\\
			& Clean &  Final & Clean &  Final &  (s) \\ \hline
			PatchBatch~\cite{Gadot2016PatchBatch} & -& -&5.79 &6.78 &50.0\\
			EpicFlow~\cite{EpicFlow} & - & - &4.12  & 6.29  &15.0\\  
			CPM-flow~\cite{Hu2016Efficient} & - & -&3.56  & 5.96 &4.30\\ 
			FullFlow~\cite{Chen2016Full} &- & 3.60 & \ud{2.71} &5.90 & 240\\
			FlowFields~\cite{Bailer2015Flow} & -& -& 3.75  & 5.81  & 28.0 \\
			MRFlow~\cite{Wulff2017Optical} & 1.83 & 3.59  &  \bd{2.53} & 5.38 & 480\\
			FlowFieldsCNN ~\cite{Bailer_2017_CVPR} & - & - & 3.78 & 5.36 &23.0\\
			DCFlow~\cite{Xu2017Accurate}& - & - & 3.54 & \ud{5.12} & 8.60\\
			SpyNet-ft~\cite{Ranjan:2016:SpyNet} & (3.17) & (4.32) & 6.64 & 8.36  &0.16\\
			FlowNet2~\cite{Ilg:2016:Flownet2} & 2.02 &   3.14	 &   3.96 & 6.02 &0.12\\
			FlowNet2-ft~\cite{Ilg:2016:Flownet2} & ({1.45})	&  ({2.01}) &  4.16 & 5.74 &0.12	\\ 
			LiteFlowNet-CVPR  & (1.64) & (2.23) & 4.86 & 6.09 &0.09 \\
			LiteFlowNet-arXiv & (\bd{1.35}) & (\bd{1.78}) & 4.54 & 5.38 &0.09\\ \hline
FlowNetS+ & 	(2.80) &	(2.76)	&  6.49 & 6.54 & \bd{0.01}\\
FlowNetC+ & 2.31	 & 2.34   & 5.04  &5.47 &0.05 \\
			PWC-Net-small & 2.83 &	4.08 & - & - &{0.02}\\
			PWC-Net-small-ft & (2.27) &	(2.45) & 5.05 & 5.32 &{0.02}\\			
			PWC-Net & 2.55 &	3.93 & - & - &{0.03}\\
			PWC-Net-ft & ({1.70})	& ({2.21}) & {3.86}  & 5.13    & {0.03}  \\	
			PWC-Net-ft-final  & ({2.02}) &({2.08})  & 4.39 & {5.04}  & {0.03}\\
			PWC-Net\_ROB  & ({1.81}) &({2.29})  & 3.90 & {4.90}  & {0.03}\\
			PWC-Net+  & (1.71 ) &(2.34)  & 3.45 & \bd{4.60}  & {0.03}\\
		\end{tabular}
	\end{table}
	
	\begin{table}[!t]
		\caption{Detailed results on the Sintel benchmark for different regions,  velocities ($s$), and distances from motion boundaries ($d$).}
		\label{tab:result:sintel:comparison} 	\vspace{-2mm}
		\footnotesize
		\centering
		\setlength\tabcolsep{1pt} 
		\begin{tabular}{lcccccccccccc} \\
			Final  & 	 matched	& unmatched& $d_{0\!-\!10}$ &$d_{10\!-\!60}$	&$d_{60\!-\!140}$&	$s_{0\!-\!10}$	&$s_{10\!-\!40}$	&$_{s40\!+\!}$\\ \hline
			PWC-Net &\bd{2.44}&\bd{27.08}&\bd{4.68}&\bd{2.08}&\bd{1.52}&\bd{0.90}&\bd{2.99}&\bd{31.28}\\
			FlowNet2 &2.75&30.11&4.82&2.56&1.74&0.96&3.23&35.54\\
			SpyNet &4.51&39.69&6.69&4.37&3.29&1.40&5.53&49.71\\
			\hline Clean \\
			PWC-Net &\bd{1.45}&\bd{23.47}&3.83&\bd{1.31}&\bd{0.56}&0.70&2.19&\bd{23.56}\\
			FlowNet2 &1.56&25.40&\bd{3.27}&1.46&0.86&\bd{0.60}&\bd{1.89}&27.35\\
			SpyNet &3.01&36.19&5.50&3.12&1.72&0.83&3.34&43.44\\
		\end{tabular}
	\end{table}

	\begin{table}[h]
		\caption{Results on the KITTI dataset. ``-ft'' means fine-tuning on the KITTI \emph{training} set and the numbers in the parenthesis are results on the data the methods have been fine-tuned on.} 	\vspace{-1mm}
		\label{tab:kitti}
		\footnotesize
		\centering
		\setlength\tabcolsep{2pt} 
		\begin{tabular}{lcccccc} \\
			\multirow{3}*{Methods} & \multicolumn{3}{c}{ KITTI 2012} & \multicolumn{3}{c}{ KITTI 2015}  \\
			& AEPE & AEPE & Fl-Noc &  AEPE   & Fl-all & Fl-all \\
			& \emph{train} & \emph{test} & \emph{test}   & \emph{train} & \emph{train}  & \emph{test} \\  \hline
			EpicFlow~\cite{EpicFlow}  & - & 3.8& 7.88\% & -& - & 26.29 \%  \\ 
			FullFlow~\cite{Chen2016Full}  & - & - & -& -& - & 23.37 \% \\ 
			CPM-flow~\cite{Hu2016Efficient} & - & 3.2 &  5.79\%& -& - & 22.40 \% \\ 
			PatchBatch~\cite{Gadot2016PatchBatch} &- &3.3&  5.29\% & -&- & 21.07\% \\ 
			FlowFields~\cite{Bailer2015Flow} & -& -& -& - &-  & 19.80\%  \\
			MRFlow~\cite{Wulff2017Optical} & - &  -&- &-  &  14.09 \% & 12.19 \% \\ 
			DCFlow~\cite{Xu2017Accurate} & - &  -&- &- & 15.09 \% &14.83 \% \\ 
			SDF~\cite{Bai2016Exploiting} & - &2.3 & {3.80}\% & - &- & 11.01 \% \\ 
			MirrorFlow~\cite{Hur_2017_ICCV} &- & 2.6 & 4.38\% & - & 9.93\% & \ud{10.29}\% \\ 
			SpyNet-ft~\cite{Ranjan:2016:SpyNet} & (4.13)  & 4.7 &12.31\% & - & - & 35.07\%  \\ 
			FlowNet2~\cite{Ilg:2016:Flownet2}  &4.09& - & - & 10.06 & 30.37\% & - \\ 
			FlowNet2-ft~\cite{Ilg:2016:Flownet2}  & ({1.28}) & {1.8} & 4.82\% & (\ud{2.30}) & ({8.61}\%) & {10.41} \% \\ 
			LiteFlowNet-CVPR  & ({1.26}) & {1.7} & - & ({2.16}) & ({8.16}\%) & {10.24} \% \\ 
			LiteFlowNet-arXiv & (\bd{1.05}) & {1.6} & \bd{3.27}\% & ({1.62}) & (\bd{5.58}\%) & {9.38} \% \\  \hline						
			PWC-Net & 4.14 & - &- & 10.35 & 33.67\% &-\\ 
			PWC-Net-ft-CVPR& (\ud{1.45})	&  {1.7}	 &4.22\%	& ({2.16})	& ({9.80}\%) & {9.60}\%  		\\  
			PWC-Net-ft  & (\ud{1.08})	&  \bd{1.5}	 &3.41\% 	& (\bd{1.45})	& (\ud{7.59}\%) & \bd{7.90}\%  		\\  			
		\end{tabular}
	\end{table}

%

\begin{figure*}[t]
	\footnotesize
	\begin{center}
		\newcommand{\figwidth}{0.18\linewidth}
		\newcommand{\Figwidth}{0.32\linewidth}
		\newcommand{\shiftfigure}{\hspace{0mm}}
		\begin{tabular}{cccc}	
			\shiftfigure \hspace{-1em}\subfigimg[width = \Figwidth]{First frame (training)}{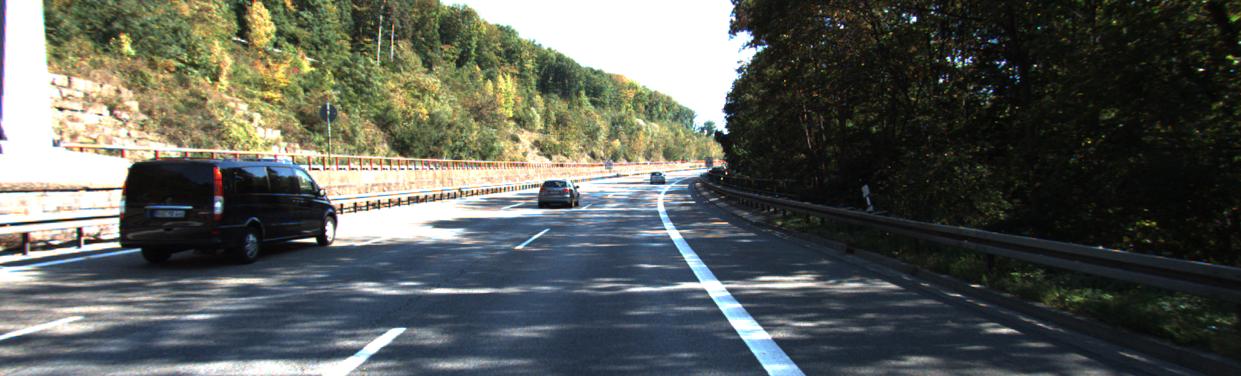} &
			\shiftfigure \hspace{-1em}\subfigimg[width = \Figwidth]{Ground truth}{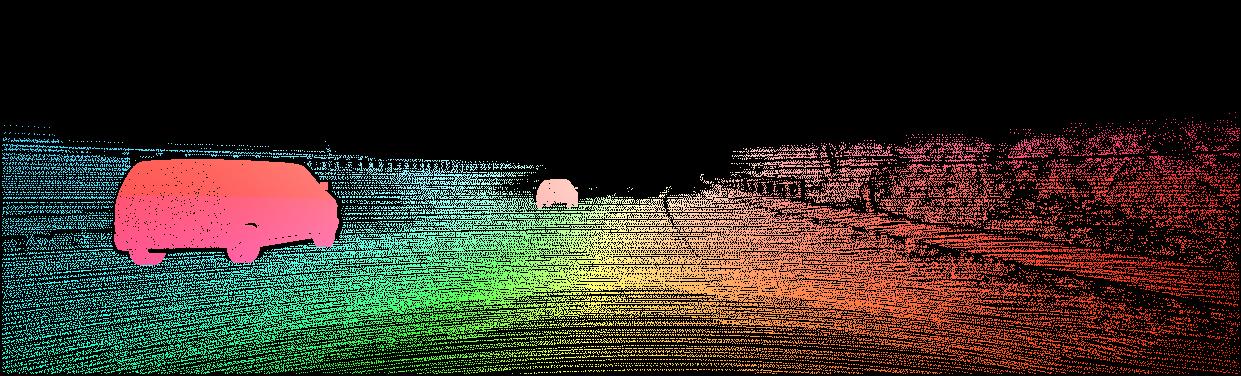} &
			\shiftfigure \hspace{-1em}\subfigimg[width = \Figwidth]{W/o context}{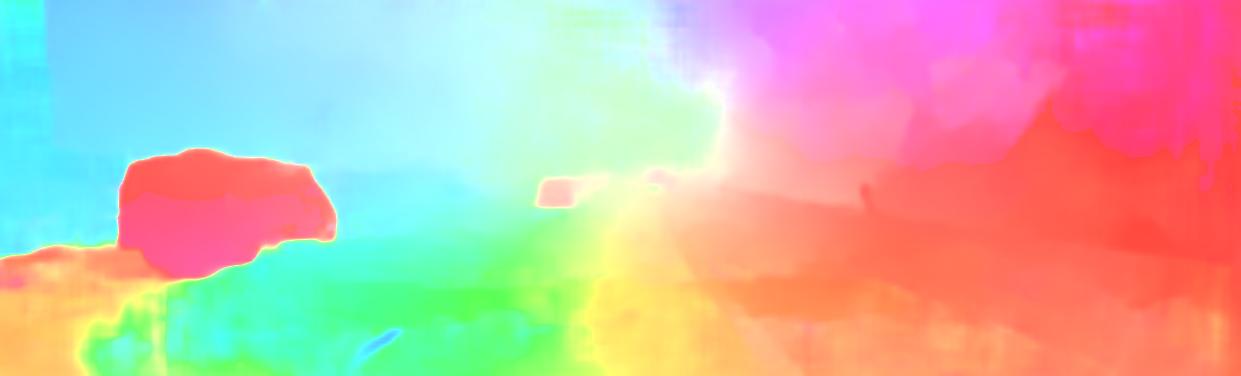} \\
			\shiftfigure \hspace{-1em}\subfigimg[width = \Figwidth]{W/o DenseNet }{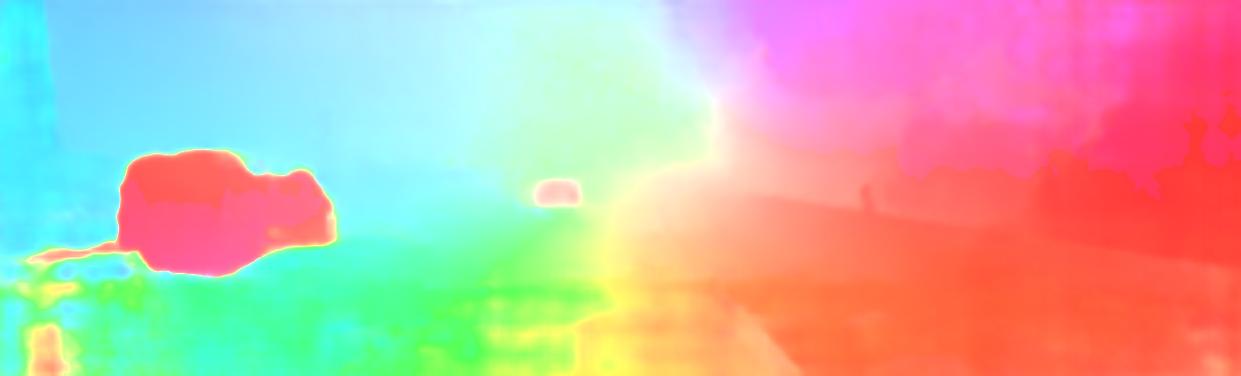} &
			\shiftfigure \hspace{-1em}\subfigimg[width = \Figwidth]{PWC-Net}{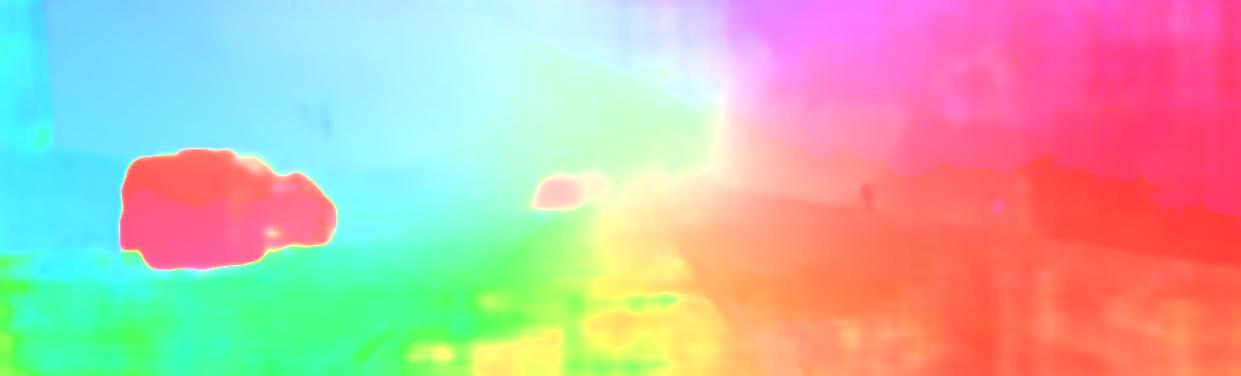} &
			\shiftfigure \hspace{-1em}\subfigimg[width = \Figwidth]{PWC-Net-KITTI-ft}{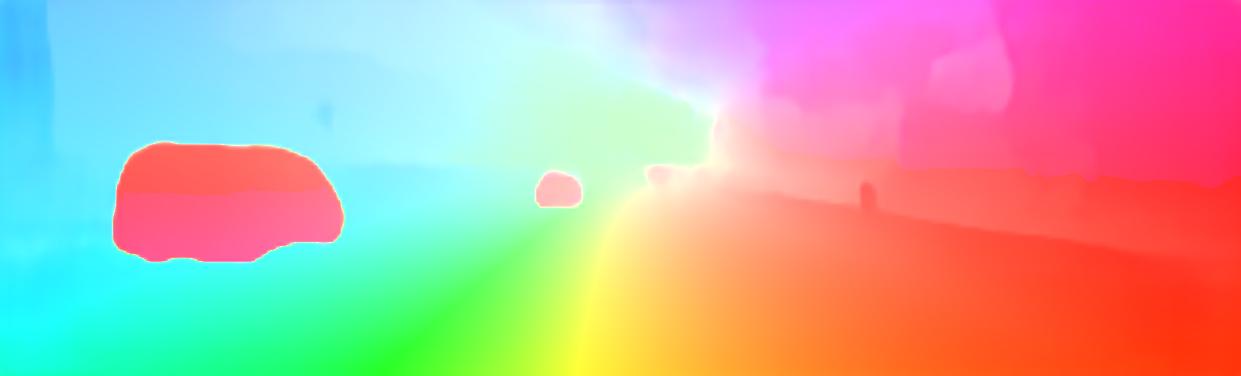}  \\			
			\shiftfigure \hspace{-1em}\subfigimg[width = \Figwidth]{First frame (test)}{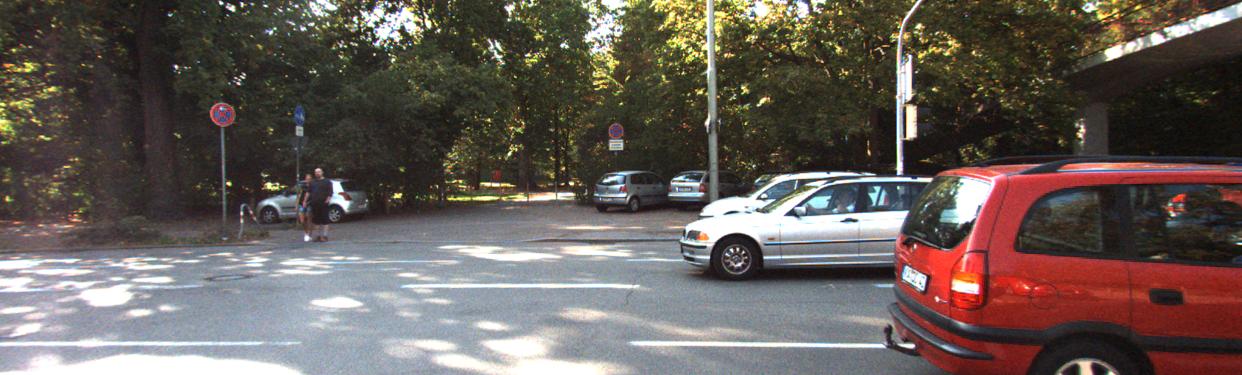} &
			\shiftfigure \hspace{-1em}\subfigimg[width = \Figwidth]{Second frame}{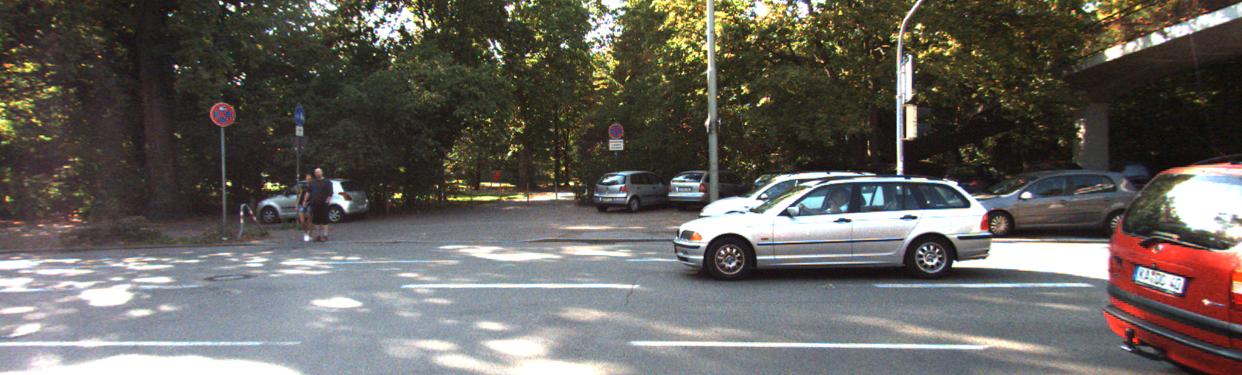} &
			\shiftfigure \hspace{-1em}\subfigimg[width = \Figwidth]{W/o context}{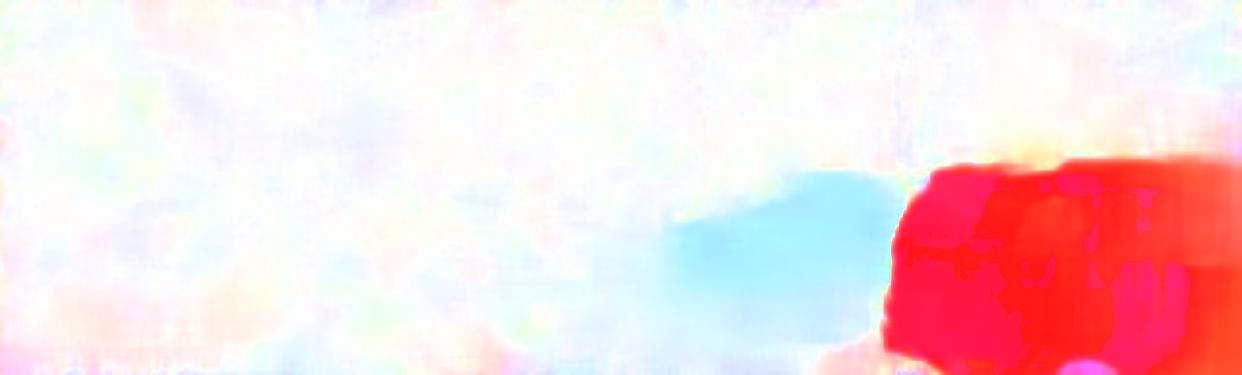} \\
			\shiftfigure \hspace{-1em}\subfigimg[width = \Figwidth]{W/o DenseNet }{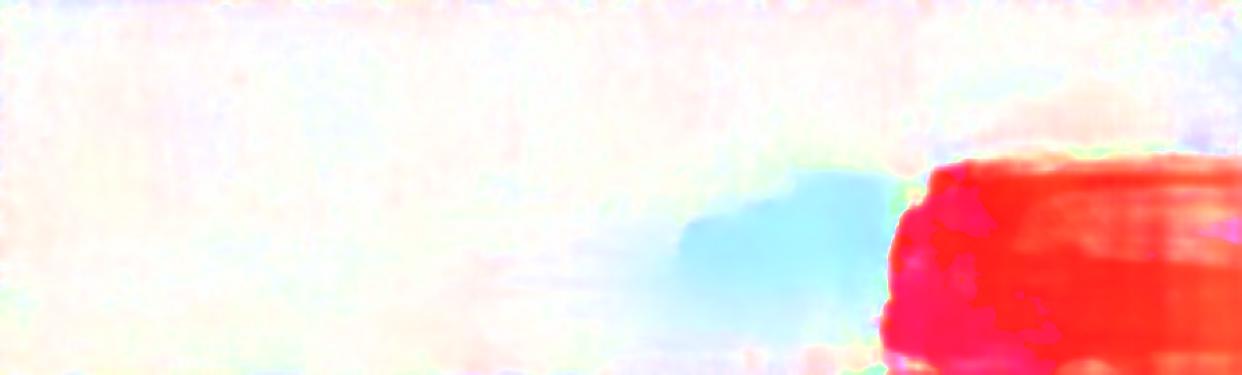} &
			\shiftfigure \hspace{-1em}\subfigimg[width = \Figwidth]{PWC-Net}{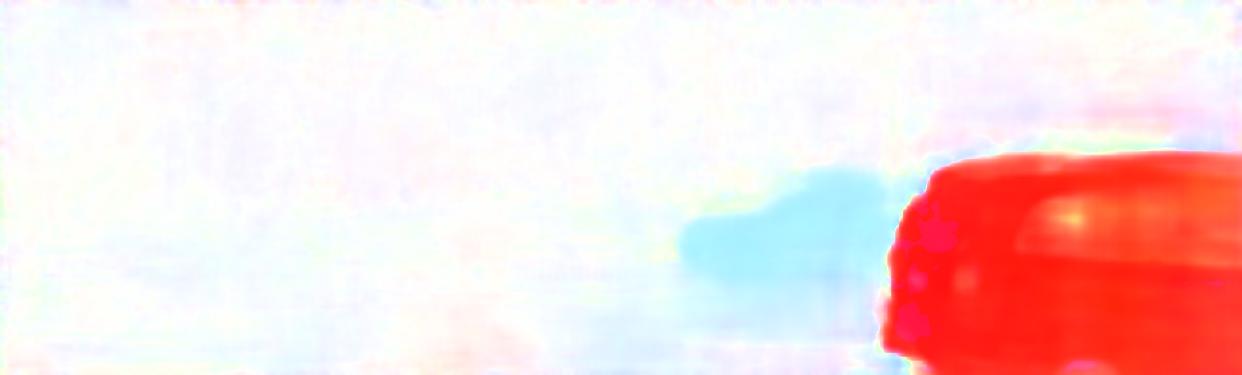} &
			\shiftfigure \hspace{-1em}\subfigimg[width = \Figwidth]{PWC-Net-KITTI-ft}{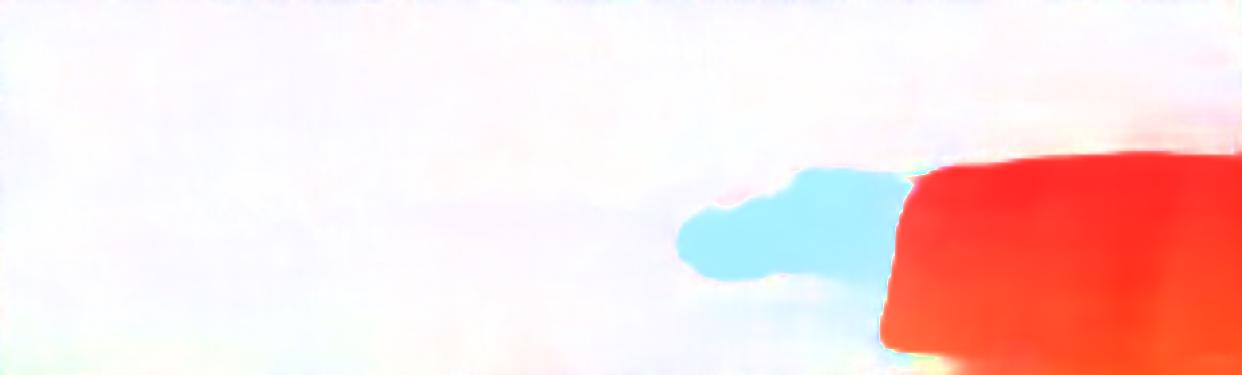}  \\							
		\end{tabular}		
	\end{center}
	\vspace{-4mm}
	\caption{Results on KITTI 2015 \emph{training} and \emph{test} sets. Fine-tuning fixes large regions of errors and recovers sharp motion boundaries. }
	\label{fig:kitti:models}
\end{figure*}

	\beforePara
	\subsubsection{KITTI.}
	When fine-tuning on KITTI, we crop $896\times 320$ image patches and reduce the amount of rotation, zoom, and squeeze during data augmentation. The batch size is 4 too. 
	The large patches can capture the large motion in the KITTI dataset. Since the ground truth is semi-dense, we upsample the predicted flow at the quarter resolution to compare with the scaled ground truth at the full resolution. We exclude the invalid pixels in computing the loss function. 
	
	The CVPR version of PWC-Net outperforms many recent two-frame optical flow methods on the 2015 set, as shown in Table~\ref{tab:kitti}. 
	On the 2012 set, PWC-Net is inferior to SDF that assumes a rigidity constraint for the background. Although the rigidity assumption works well on the static scenes in the 2012 set, PWC-Net outperforms SDF in the 2015 set which mainly consists of dynamic scenes and is more challenging. The visual results in Fig.~\ref{fig:kitti:models} qualitatively demonstrate the benefits of using the context network, DenseNet connections, and fine-tuning, respectively. In particular, fine-tuning fixes large regions of errors in the test set, demonstrating the benefit of learning when the training and test data share similar statistics.

		\begin{figure}[h]
		\footnotesize
		\begin{center}
			\newcommand{\figwidth}{0.18\linewidth}
			\newcommand{\Figwidth}{0.48\linewidth}
			\newcommand{\shiftfigure}{\hspace{1mm}}
			\shiftfigure \subfigimg[width = \Figwidth]{First frame }{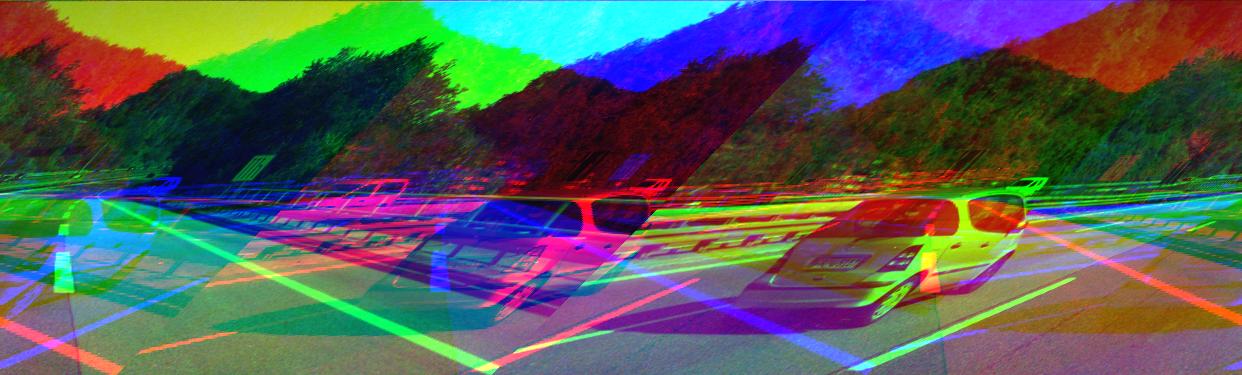} 
			\shiftfigure \subfigimg[width = \Figwidth]{Ground truth}{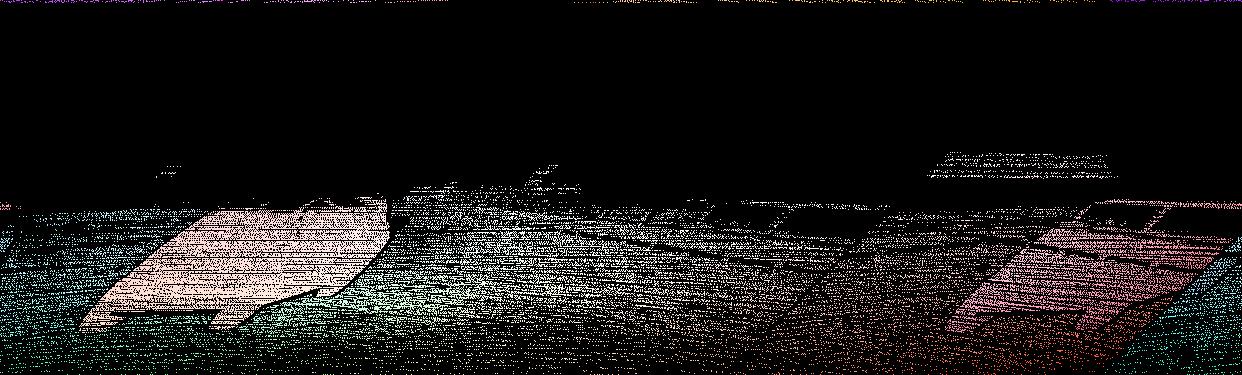}  							
		\end{center}
		\vspace{-4mm}
		\caption{Improperly read images and flow fields due to an I/O bug. }
		\label{fig:kitti:bug}
	\end{figure}
	
	\textbf{An I/O bug.} We use the Caffe code\cite{Dosovitskiy:2015Flownet} to make all the image pairs and flow fields into a single LMDB file for training. The code requires that all the input images are of the same resolution. The size of the first 156 sequences of KITTI 2015 is $375\times 1242$, but the last $44$ are of different resolutions, including $370\times1224$, $374\times1238$, and $376\times1241$. The Caffe code cannot read the last $44$ sequences properly, as shown in Fig.~\ref{fig:kitti:bug}. As a results, PWC-Net has been trained using only 156 ``good'' sequences and 44 ``bad'' ones. As a remedy, we crop all the sequences to the size of $370 \times 1224$, because there is no reliable way to resize the sparse ground truth. Re-training with the correct 200 sequences leads to about 20\% improvement on the test set of KITTI 2012 (Fl-Noc 4.22\% $\rightarrow$ 3.41\%) and 2015 (Fl-all 9.60\%  $\rightarrow$ 7.90\%). At the time of writing, PWC-Net is ranked second in non-occluded regions among all methods on KITTI 2015. It is surpassed by only one recent scene flow method that uses stereo input and semantic information~\cite{Behl2017ICCV} (Fl-all scores: 5.07\% vs 4.69\%) and more accurate than other scene flow methods, such as another recent one that also uses semantic information~\cite{ren2017cascaded}. Note that scene flow methods can use the estimated depth and the camera motion to predict the flow of out-of-boundary pixels and thus tend to have better accuracy in all regions.

	\begin{table*}[t]
		\caption{\bd{Ablation experiments}. Unless explicitly stated, the models have been trained on the FlyingChairs dataset.  }
		\label{tab:result:ablation}			
		\newcommand{\colwidth}{20}
		\subfloat[Larger-capacity \bd{feature pyramid extractor} has better performance. Learning features leads to significantly better results than fixed image pyramids. \label{tab:ablation:encoder}]{
			\tablestyle{4pt}{1.0}
			\begin{tabular}{l|x{\colwidth}x{\colwidth}x{\colwidth}x{\colwidth}x{\colwidth}x{\colwidth}x{\colwidth}} 
				& \multirow{2}*{Chairs} & Sintel & Sintel & \multicolumn{2}{c}{KITTI 2012} & \multicolumn{2}{c}{KITTI 2015} \\[.1em]
				&  & Clean & Final &  AEPE & Fl-all & AEPE & Fl-all \\[.1em]		
				\hline 
				Full model &2.00&3.33&4.59&5.14&28.67\%&13.20&41.79\%\\
				Feature $\uparrow$ &\bd{1.92}&\bd{3.03}&\bd{4.17}&\bd{4.57}&\bd{26.73}\%&\bd{11.64}&\bd{39.80}\%\\
				Feature  $\downarrow$ &2.18&3.36&4.56&5.75&30.79\%&14.05&44.92\%\\
				Image & 2.95&4.42&5.58&7.28&31.25\%&16.29&45.13\%\\	
		\end{tabular}}\hspace{5mm}
		\subfloat[\bd{Cost volume.} Removing the cost volume ($0$) results in moderate performance loss. PWC-Net can handle large motion using a small search range to compute the cost volume. \label{tab:ablation:cv}]{
			\tablestyle{4pt}{1.0}
			\begin{tabular}{c|x{\colwidth}x{\colwidth}x{\colwidth}x{\colwidth}x{\colwidth}x{\colwidth}x{\colwidth}} 
				Max. & \multirow{2}*{Chairs} & Sintel & Sintel & \multicolumn{2}{c}{KITTI 2012} & \multicolumn{2}{c}{KITTI 2015} \\[.1em]
				Disp. &  & Clean & Final &  AEPE & Fl-all & AEPE & Fl-all \\[.1em]		
				\hline 
				$0$ &2.13 &3.66&5.09&5.25&29.82\%&13.85&43.52\%\\
				$2$ &2.09&\bd{3.30}&\bd{4.50}&5.26&\bd{25.99}\%&13.67&\bd{38.99}\%\\
				Full model ($4$) &2.00&3.33&4.59&5.14&28.67\%&13.20&41.79\%\\
				$6$ &\bd{1.97}&3.31&4.60&\bd{4.96}&27.05\%&\bd{12.97}&40.94\%\\
		\end{tabular}}\hspace{5mm}	
		\vspace{-1mm}		
		\subfloat[\bd{More feature pyramid levels} help after fine-tuning on FlyingThings.  \label{tab:ablation:feature:level}]{
			\tablestyle{4pt}{1.0}
			\begin{tabular}{l|x{\colwidth}x{\colwidth}x{\colwidth}x{\colwidth}x{\colwidth}x{\colwidth}x{\colwidth}} 
				& \multicolumn{3}{c}{Trained on FlyingChairs}  & \multicolumn{3}{c}{Fine-tuned on FlyingThings}  \\[.1em]
				& Chairs  & Clean & Final &  Chairs & Clean & Final \\[.1em]		
				\hline 
5-level &2.13&3.28&4.52&2.62	&2.98	&4.29	\\ 
6-level &\bd{1.95}&\bd{2.96}&\bd{4.32}&\bd{2.28} &	\bd{2.50}&3.97\\
Full model (7) &{2.00}&3.33&4.59&{2.30}&2.55&\bd{3.93}\\
		\end{tabular}}\hspace{5mm}			
		\subfloat[Larger-capacity \bd{optical flow estimator} has better performance. \label{tab:ablation:estimator}]{
			\tablestyle{4pt}{1.0}
			\begin{tabular}{l|x{\colwidth}x{\colwidth}x{\colwidth}x{\colwidth}x{\colwidth}x{\colwidth}x{\colwidth}} 
				& \multirow{2}*{Chairs} & Sintel & Sintel & \multicolumn{2}{c}{KITTI 2012} & \multicolumn{2}{c}{KITTI 2015} \\[.1em]
				&  & Clean & Final &  AEPE & Fl-all & AEPE & Fl-all \\[.1em]		
				\hline 
				Full model &2.00&3.33&4.59&5.14&28.67\%&13.20&41.79\%\\
				Estimator $\uparrow$ &\bd{1.92}&\bd{3.09}&\bd{4.50}&\bd{4.64}&\bd{25.34}\%&\bd{12.25}&\bd{39.18}\%\\
				Estimator  $\downarrow$ &2.01&3.37&4.58&4.82&26.35\%&12.83&40.53\%\\
		\end{tabular}}\hspace{5mm}
		\vspace{-1mm}		
		\subfloat[\bd{Context network} consistently helps; \bd{DenseNet} helps after fine-tuning on FlyingThings.  \label{tab:ablation:context}]{
			\tablestyle{4pt}{1.0}
			\begin{tabular}{l|x{\colwidth}x{\colwidth}x{\colwidth}x{\colwidth}x{\colwidth}x{\colwidth}x{\colwidth}} 
				& \multicolumn{3}{c}{Trained on FlyingChairs}  & \multicolumn{3}{c}{Fine-tuned on FlyingThings}  \\[.1em]
				& Chairs  & Clean & Final &  Chairs & Clean & Final \\[.1em]		
				\hline 
				Full model &\bd{2.00}&3.33&4.59&\bd{2.34}&\bd{2.60}&\bd{3.95}\\
				No DenseNet &2.06&\bd{3.09}&\bd{4.37}&2.48&2.83&4.08\\
				No Context &2.23&3.47&4.74&2.55&2.75&4.13\\
		\end{tabular}}\hspace{5mm}
		\subfloat[Independent runs with different initializations lead to minor performance differences.  \label{tab:ablation:ind:runs}]{
			\tablestyle{4pt}{1.0}
			\begin{tabular}{l|x{\colwidth}x{\colwidth}x{\colwidth}x{\colwidth}x{\colwidth}x{\colwidth}x{\colwidth}} 
				& \multirow{2}*{Chairs} & Sintel & Sintel & \multicolumn{2}{c}{KITTI 2012} & \multicolumn{2}{c}{KITTI 2015} \\[.1em]
				&  & Clean & Final &  AEPE & Fl-all & AEPE & Fl-all \\[.1em]		
				\hline 
            1st run &2.00&3.33&4.59&5.14&28.67\%&13.20&41.79\%\\
            2nd run &{2.00}&\bd{3.23}&\bd{4.36}&\bd{4.70}&\bd{25.52}\%&\bd{12.57}&\bd{39.06}\%\\
            3rd run &2.00&3.33&4.65&4.81&27.12\%&13.10&40.84\%\\
		\end{tabular}}\hspace{5mm}
		\subfloat[\bd{Warping layer} is a critical component for the performance.  \label{tab:ablation:warping}]{
			\tablestyle{4pt}{1.0}
			\begin{tabular}{l|x{\colwidth}x{\colwidth}x{\colwidth}x{\colwidth}x{\colwidth}x{\colwidth}x{\colwidth}} 
				& \multirow{2}*{Chairs} & Sintel & Sintel & \multicolumn{2}{c}{KITTI 2012} & \multicolumn{2}{c}{KITTI 2015} \\[.1em]
				&  & Clean & Final &  AEPE & Fl-all & AEPE & Fl-all \\[.1em]		
				\hline 
				Full model &\bd{2.00}&\bd{3.33}&\bd{4.59}&\bd{5.14}&\bd{28.67}\%&\bd{13.20}&\bd{41.79}\%\\
				No warping &2.17&3.79&5.30&5.80&32.73\%&13.74&44.87\%\\
		\end{tabular}}\hspace{5mm}
		\subfloat[\bd{Residual connections} in the optical flow estimator are helpful.  \label{tab:ablation:residual}]{
			\tablestyle{4pt}{1.0}
			\begin{tabular}{l|x{\colwidth}x{\colwidth}x{\colwidth}x{\colwidth}x{\colwidth}x{\colwidth}x{\colwidth}} 
				& \multirow{2}*{Chairs} & Sintel & Sintel & \multicolumn{2}{c}{KITTI 2012} & \multicolumn{2}{c}{KITTI 2015} \\[.1em]
				&  & Clean & Final &  AEPE & Fl-all & AEPE & Fl-all \\[.1em]		
				\hline 
				Full model &2.00&3.33&4.59&5.14&28.67\%&13.20&41.79\%\\
				Residual &\bd{1.96}&\bd{3.14}&\bd{4.43}&\bd{4.87}&\bd{27.74}\%&\bd{12.58}&\bd{41.16}\%\\
		\end{tabular}}\hspace{5mm}
	\end{table*}
	
	\subsubsection[Robust Vision Challenge]{Robust Vision Challenge\footnote{ \url{http://www.robustvision.net} }}
	
\begin{figure}[h]
		\begin{center}	
			\includegraphics[width=0.95\linewidth]{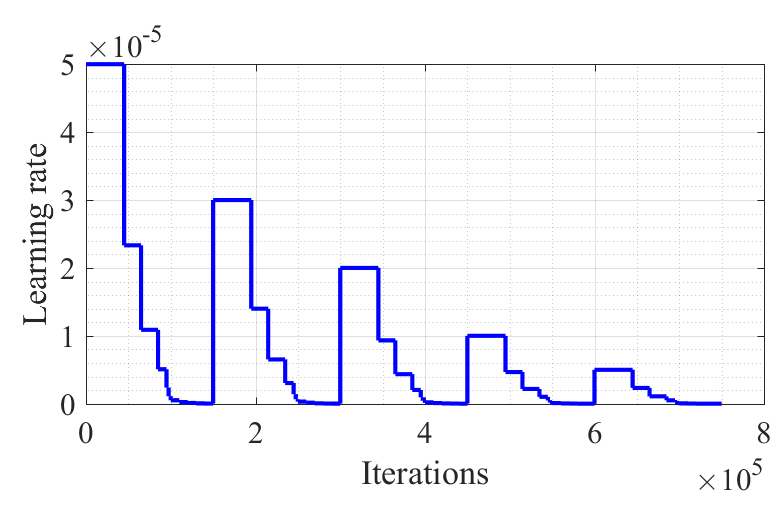} 
		\end{center}
		\caption{Learning rate schedule for fine-tuning using data from Sintel, KITTI, and HD1K. For this mixed dataset, we use more iterations and learning rate disruptions than the learning rate schedule in Fig.~\ref{fig:ft:lr}.}
		\label{fig:rob:lr}
	\end{figure}

	PWC-Net\_ROB is the winning entry in the optical flow competition of the robust vision challenge, which requires applying a method using the same parameter setting to four benchmarks: Sintel~\cite{Butler:ECCV:2012}, KITTI 2015~\cite{Menze2015CVPR}, HD1K~\cite{kondermann2016hci}, and Middlebury~\cite{Baker:2011:DEO}. 
	To participate the challenge, we fine-tune the model using training data from Sintel, KITTI 2015, and HD1K and name it as PWC-Net\_ROB. We do not use the Middlebury training data because the provided eight image pairs are too small and of low resolution compared to other datasets. We use a batch size of $6$, with $2$ image pairs from Sintel, KITTI, and HD1K respectively. 
   The cropping size is $768\times320$ for Sintel and KITTI 2015. For HD1K, we first crop $1536\times640$ patches and then downsample the cropped images and flow fields to  $768\times320$. For the mixed datasets, we use more iterations and learning rate disruptions, as shown in Fig.~\ref{fig:rob:lr}. 
	
	Using mixed datasets increases the test error on KITTI 2015 (F-all 9.60\% $\rightarrow$ 11.63\%) but reduces the test error on MPI Sintel final (AEPE 5.04\% $\rightarrow$ 4.9). There is a larger mismatch between the training and test data of Sintel than those of KITTI 2015. Thus, using more diverse datasets reduces the over-fitting errors on Sintel. We further use a batch size of $4$, with $2$ image pairs from Sintel,  $1$ from KITTI, and $1$ from HD1K respectively,  which is our \textbf{training procotol II}. It results in a further performance gain, \ie, PWC-Net+ in Table~\ref{tab:result:sintel:comparison}. 
	
	The Middlebury images are of lower resolution and we upsample them so that the larger of the width and height of the upsampled image is around 1000 pixels. PWC-Net\_ROB has similar performance as the Classic+NL method (avg. training EPE 0.24 vs. 0.22; avg. test EPE 0.33 vs 0.32). 
	
\begin{figure*}[t]
	\footnotesize
	\begin{center}
		\newcommand{\figwidth}{0.18\linewidth}
		\newcommand{\Figwidth}{0.32\linewidth}
		\newcommand{\shiftfigure}{\hspace{0mm}}
		\begin{tabular}{cccc}				
			\shiftfigure \hspace{-1em}\subfigimg[width = \Figwidth]{First frame of Teddy }{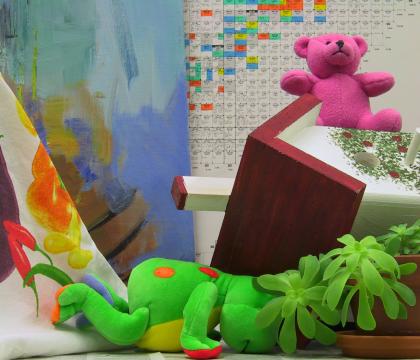} &
			\shiftfigure \hspace{-1em}\subfigimg[width = \Figwidth]{Second frame}{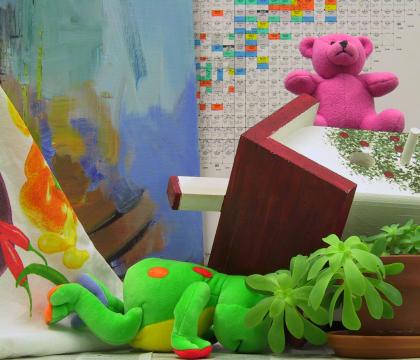} &
			\shiftfigure \hspace{-1em}\subfigimg[width = \Figwidth]{PWC-Net\_ROB}{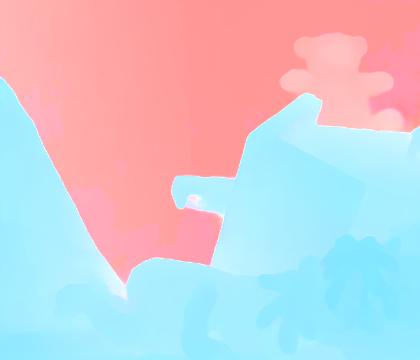} \\						
			\shiftfigure \hspace{-1em}\subfigimg[width = \Figwidth]{First frame of Grove }{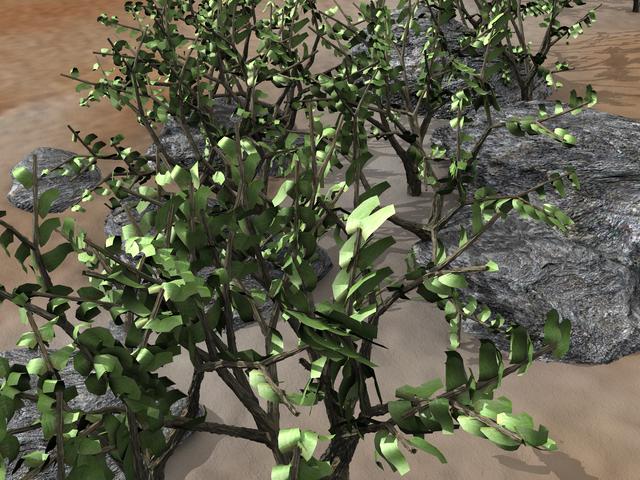} &
			\shiftfigure \hspace{-1em}\subfigimg[width = \Figwidth]{Second frame}{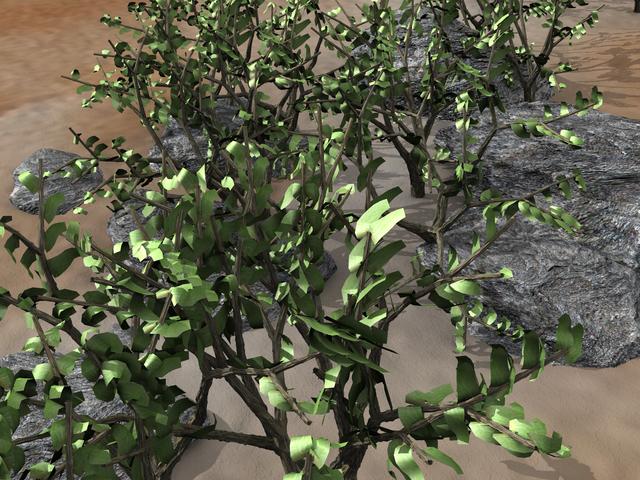} &
			\shiftfigure \hspace{-1em}\subfigimg[width = \Figwidth]{PWC-Net\_ROB}{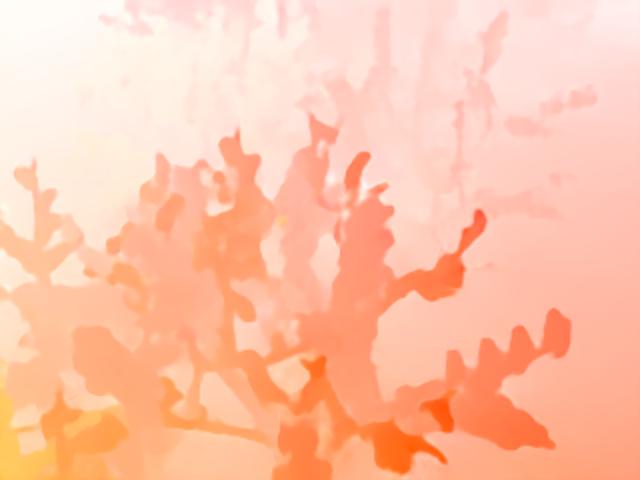} \\				
\shiftfigure \hspace{-1em}\subfigimg[width = \Figwidth]{First frame of HD1K sequence 02 }{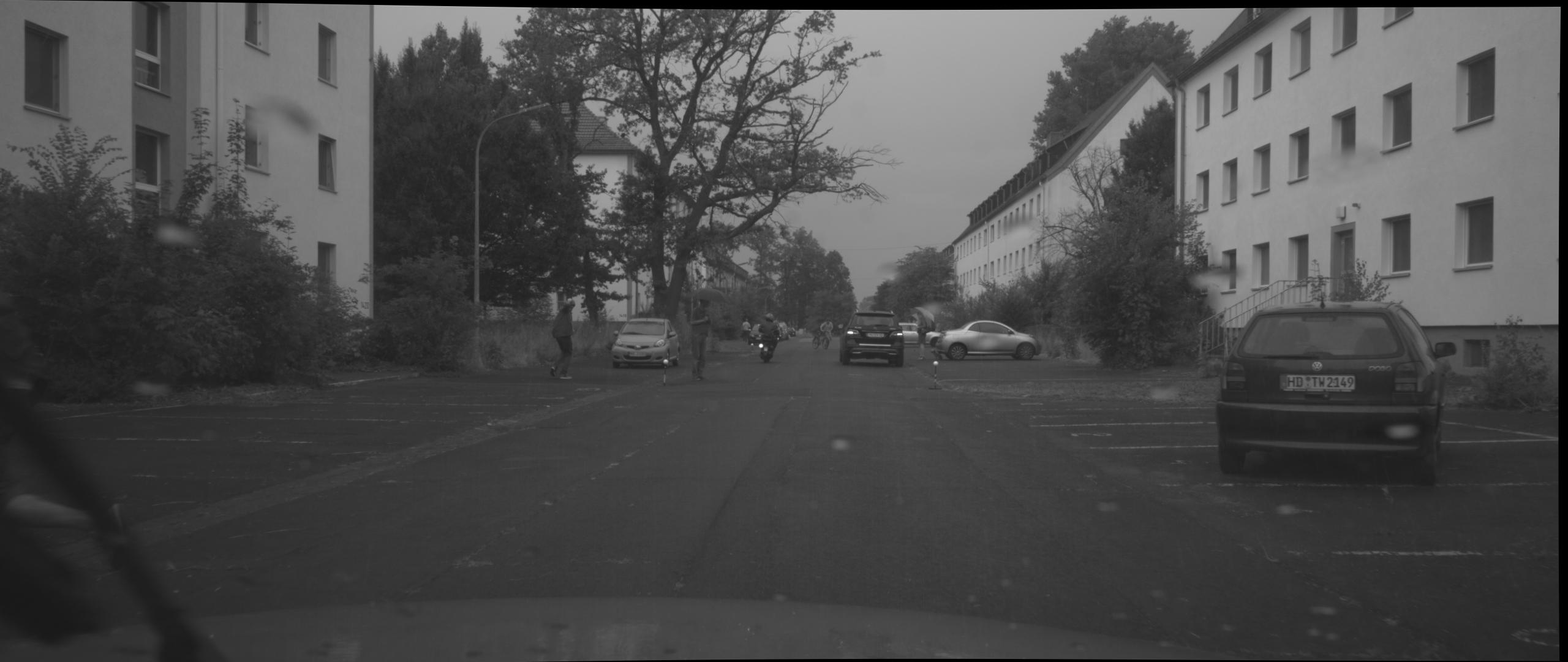} &
\shiftfigure \hspace{-1em}\subfigimg[width = \Figwidth]{Second frame}{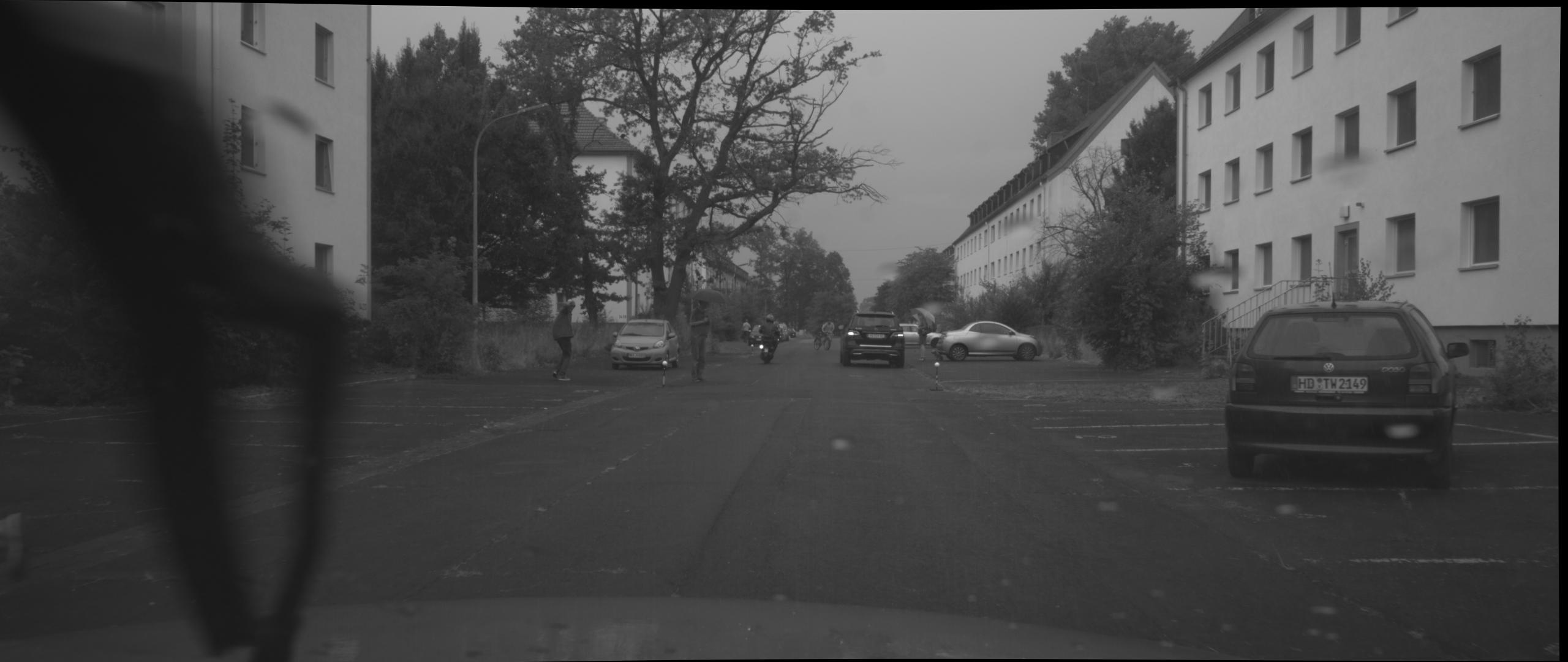} &
\shiftfigure \hspace{-1em}\subfigimg[width = \Figwidth]{PWC-Net\_ROB}{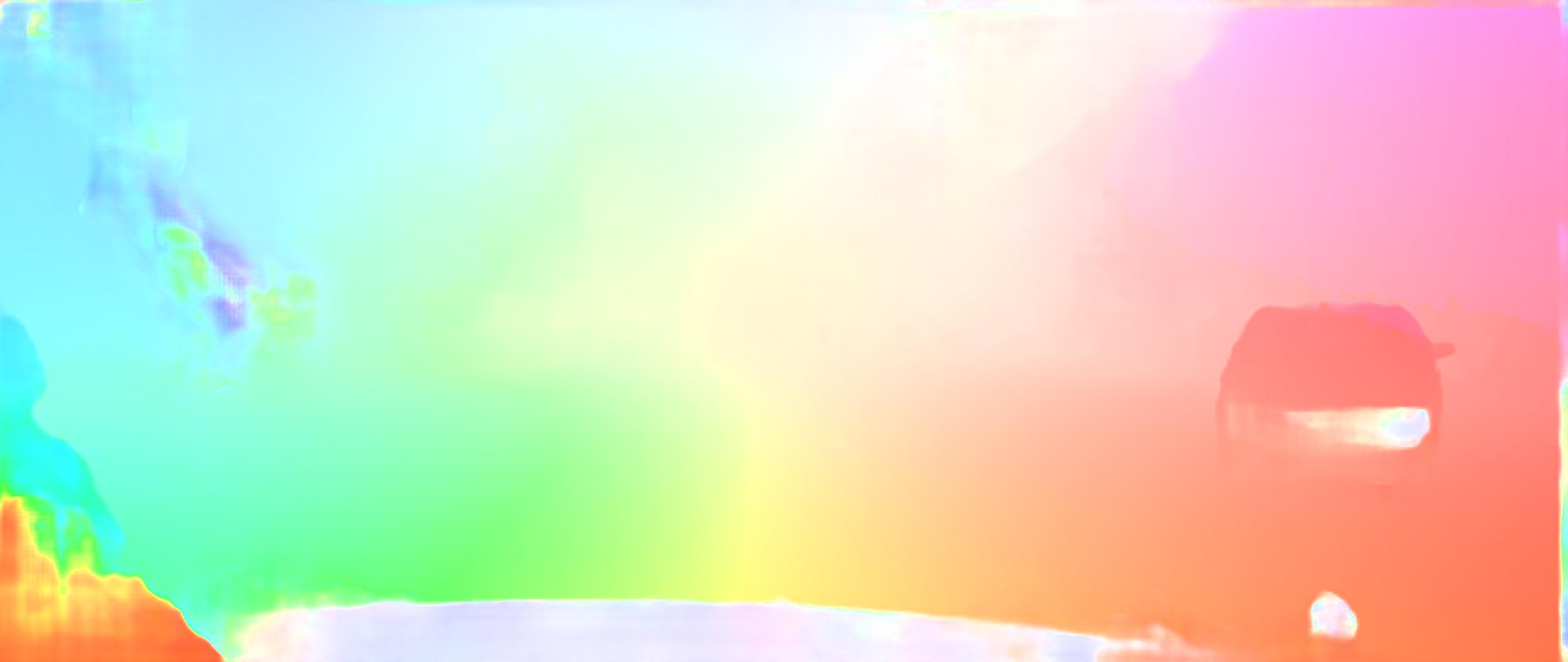} \\				
\shiftfigure \hspace{-1em}\subfigimg[width = \Figwidth]{First frame of  HD1K sequence 19 }{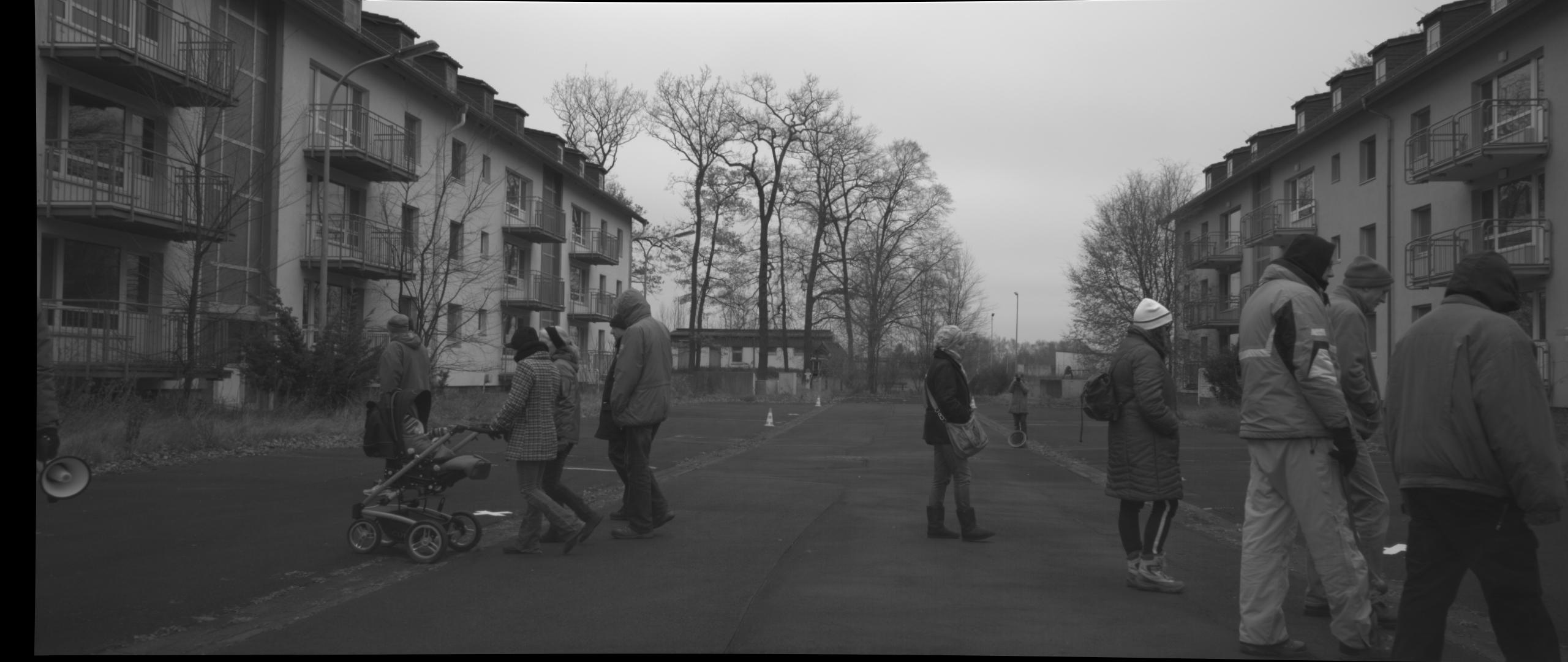} &
\shiftfigure \hspace{-1em}\subfigimg[width = \Figwidth]{Second frame}{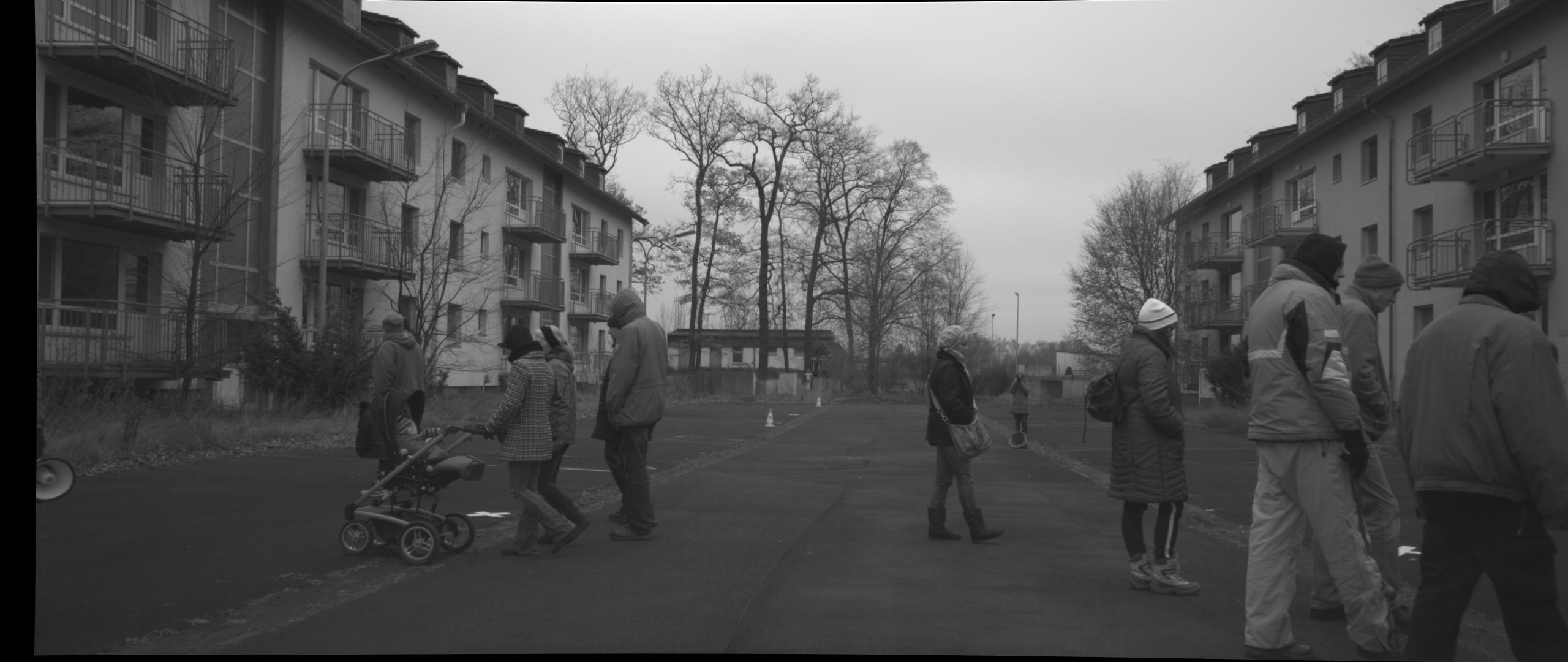} &
\shiftfigure \hspace{-1em}\subfigimg[width = \Figwidth]{PWC-Net\_ROB}{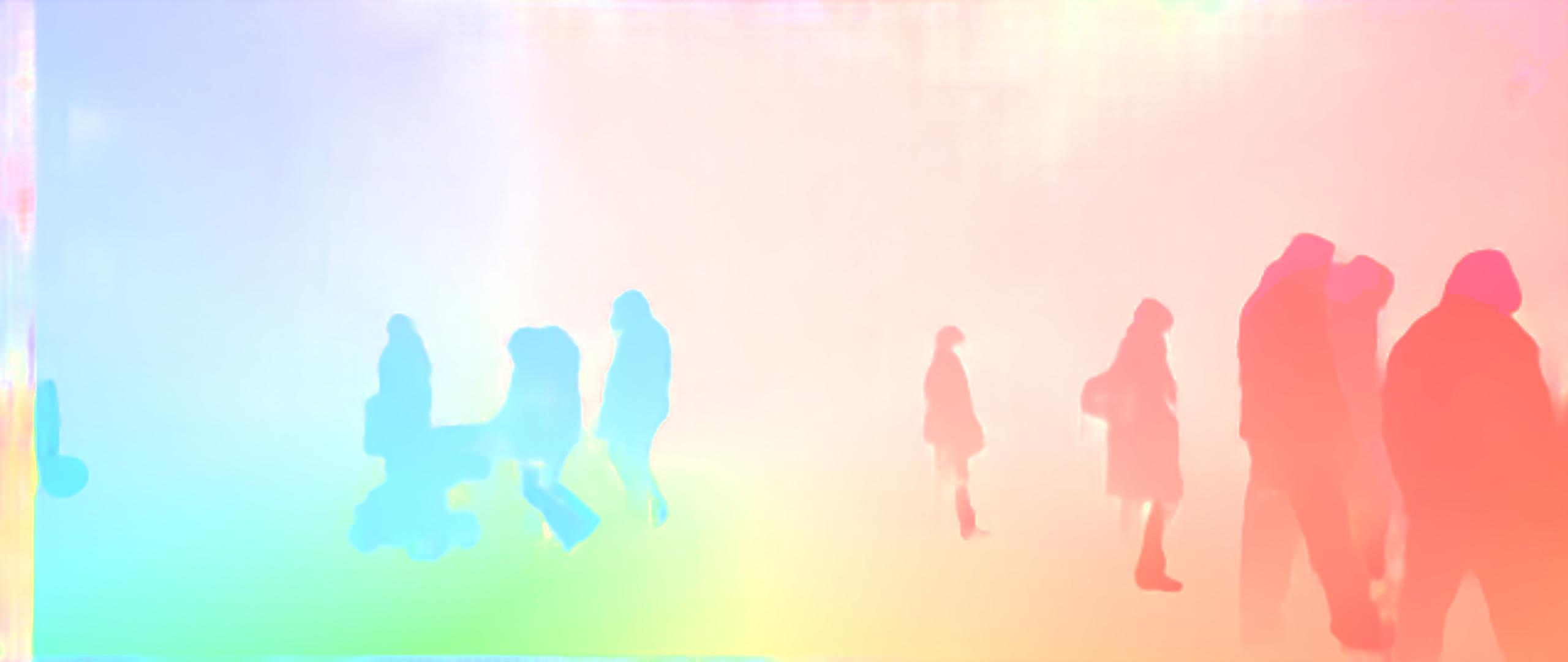} \\		
		\end{tabular}		
	\end{center}
	\vspace{-4mm}
	\caption{Results on Middlebury and HD1K test sets. PWC-Net\_ROB has not been trained using the training data of Middlebury but performs reasonably well on the test set. It cannot recover the fine motion details of the twigs in Grove though.  PWC-Net\_ROB has reasonable results in the regions occluded by the windshield wipers in sequence 02 of the HD1K test set.}
	\label{fig:MD:HD1K}
\end{figure*}

	PWC-Net\_ROB is ranked first on the HD1K benchmark~\cite{kondermann2016hci}, which consists of real-world images corrupted by rain, glare, and windshield wipers \etc. The $2560\times1080$ resolution images causes out-of-memory issue on an NVIDIA Pascal TitanX GPU with 12GB memory and requires an NVIDIA Volta 100 GPU with 16GB memory. Figure~\ref{fig:MD:HD1K} shows some visual results on Middlebury and HD1K test set. Despite minor artifacts, PWC-Net\_ROB performs robustly across these benchmarks using the same set of parameters. 
	
	\subsection{Ablation Experiments}
		
	\paragraph{Feature pyramid extractor.}
	PWC-Net uses a two-layer CNN to extract features at each pyramid level. Table~\ref{tab:ablation:encoder} summarizes the results of two variants that use one layer ($\downarrow$) and three layers ($\uparrow$) respectively. A larger-capacity feature pyramid extractor leads to consistently better results on both the training and validation datasets. Replacing the feature pyramids with image pyramids results in about 40\% loss in accuracy, confirming the benefits of learning features. 
	
	To further understand the effect of the pyramids, we test feature pyramids with different levels, as shown in Table~\ref{tab:ablation:feature:level}. Using 5-level pyramids leads to consistently worse results. Using 6-level pyramids has better performance than the default 7-level pyramids when trained on FlyingChairs, but the two have close performance after fine-tuning using FlyingThings3D. One possible reason is that the cropping size for FlyingChairs ($448\times384$) is too small for the 7-level pyramids. The size of the top level is $7\times6$, too small for a search range of $4$ pixels. By contrast, the cropping size for FlyingThings3D ($768\times384$) is better suited for the 7-level-pyramids.

	\beforePara
	\paragraph{Optical flow estimator.}
	PWC-Net uses a five-layer CNN in the optical flow estimator at each level. Table~\ref{tab:ablation:estimator} shows the results by two variants that use four layer ($\downarrow$) and seven layers ($\uparrow$) respectively. A larger-capacity optical flow estimator leads to better performance. However, we observe in our experiments that a deeper optical flow estimator might get stuck at poor local minima, which can be detected by checking the validation errors after a few thousand iterations and fixed by running from a different random initialization. 
	
	Removing the context network results in larger errors on both the training and validation sets (Table~\ref{tab:ablation:context}). Removing the DenseNet connections results in higher training error but lower validation errors when the model is trained on FlyingChairs. However, after the model is fine-tuned on FlyingThings3D, DenseNet leads to lower  errors. 
	
	We also test a residual version of the optical flow estimator, which estimates a flow increment and adds it to the initial flow to obtain the refined flow. As shown in Table~\ref{tab:ablation:residual}, this residual version slightly improves the performance. 

	\beforePara
	\paragraph{Cost volume.}
	We test the search range to compute the cost volume, shown in Table~\ref{tab:ablation:cv}. Removing the cost volume results in consistent worse results. A larger range leads to lower training error. However, all three settings have similar performance on Sintel, because a range of $2$ at every level can already deal with a motion up to 200 pixels at the input resolution. A larger range has lower EPE on KITTI, likely because the images from the KITTI dataset have larger displacements than those from Sintel. A smaller range, however, seems to force the network to ignore pixels with extremely large motion and focus more on small-motion pixels, thereby achieving lower Fl-all scores.

	\beforePara
	\paragraph{Warping.}
	Warping allows for estimating a small optical flow (increment) at each pyramid level to deal with a large optical flow. Removing the warping layers results in a significant loss of accuracy (Table~\ref{tab:ablation:warping}). Without the warping layer, PWC-Net still produces reasonable results, because the default search range of 4 to compute the cost volume is large enough to capture the motion of most sequences at the low-resolution pyramid levels. 
	
	\beforePara
	\paragraph{Independent Runs.}
    To test the robustness to the initializations, we train PWC-Net with different runs.  These independent runs have almost the same training error but some minor differences in performance on the validation sets, as shown in Table~\ref{tab:ablation:ind:runs}. 
	
	\beforePara
	\paragraph{Dataset scheduling.}
	We also train PWC-Net using different dataset scheduling schemes, as shown in Table~\ref{tab:result:data}. Sequentially training on FlyingChairs, FlyingThings3D, and Sintel gradually improves the performance, consistent with the observations in~\cite{Ilg:2016:Flownet2}. Directly training using the test data leads to good ``over-fitting'' results, but the trained model does not perform as well on other datasets. 
	
	\begin{table}[!t]
		\caption{ \textbf{Training dataset schedule} leads to better local minima. () indicates results on the dataset the method has been trained on. }
		\label{tab:result:data} \vspace{-2mm}
		\scriptsize
		\centering
		\setlength\tabcolsep{2pt} 
		\begin{tabular}{lccccccc} \\
			\multirow{2}*{Data}	& Chairs & \multicolumn{2}{c}{Sintel (AEPE)} & \multicolumn{2}{c}{ KITTI 2012} & \multicolumn{2}{c}{ KITTI 2015}  \\ 
			& AEPE & Clean &  Final & AEPE & Fl-all & AEPE & Fl-all \\ \hline
			Chairs & (\bd{2.00})&3.33&4.59&5.14&28.67\%&13.20&41.79\%\\
			Chairs-Things &2.30&2.55&3.93&4.14&21.38\%&10.35&33.67\%\\
			Chairs-Things-Sintel &2.56& (\bd{1.70})& (\bd{2.21})&\bd{2.94}&\bd{12.70}\%&\bd{8.15}&\bd{24.35}\%\\
			Sintel &3.69&(1.86)&(2.31)&3.68&16.65\%&10.52&30.49\%\\
		\end{tabular}
	\end{table}

	\beforePara
	\paragraph{Model size and running time.} Table~\ref{tab:time:memory} summarizes the model size for different CNN models.  PWC-Net has about 17 times fewer parameters than FlowNet2. PWC-Net-small 
	further reduces this by an additional 2 times via dropping DenseNet connections and is more suitable for memory-limited applications. 
	
	The timings have been obtained on the same desktop with an NVIDIA Pascal TitanX GPU. For more precise timing, we exclude the reading and writing time when benchmarking the forward and backward inference time. PWC-Net is about 2 times faster in forward inference and at least 3 times faster in training than FlowNet2. 
	
	\begin{table}[h]
		\caption{ \bd{Model size and running time.} PWC-Net-small drops DenseNet connections. For training, the lower bound of 14 days for FlowNet2 is obtained by 6(FlowNetC) + 2$\times$4 (FlowNetS). The inference time is for $1024\times448$ resolution images.}
		\label{tab:time:memory} \vspace{-1mm}
		\scriptsize
		\centering
		\setlength\tabcolsep{1pt} 
		\begin{tabular}{lccccccc} \\
			Methods & FlowNetS & FlowNetC & FlowNet2 & SpyNet  & PWC-Net & PWC-Net-small \\ \hline
			\#parameters (M) & 38.67 & 39.17 &  162.49&  1.2  &8.75 &  4.08  \\ 
			Parameter Ratio		& 23.80\%& 24.11\% &100\% & 0.74\% & 5.38\% & 2.51\% \\
			Memory (MB) & 154.5 & 156.4 & 638.5& 9.7  & 41.1  & 22.9 \\
			Memory Ratio &24.20\% &	24.49\% &	100\% &	1.52\%   &	6.44\% &	3.59\%\\ \hline
			Training (days) & 4  & 6 & $>$14 & -  &4.8 & 4.1 \\ 
			Forward (ms) &11.40	&21.69	&84.80 & - 		&28.56 &20.76\\
			Backward (ms)	&16.71	&48.67	&78.96& -		&44.37 &28.44 \\
		\end{tabular}
	\end{table}

\begin{figure*}[th]
	\footnotesize
	\begin{center}
		\newcommand{\figwidth}{0.18\linewidth}
		\newcommand{\Figwidth}{0.24\linewidth}
		\newcommand{\shiftfigure}{\hspace{0mm}}
		\begin{tabular}{cccc}
			\shiftfigure \hspace{-1em}\subfigimg[width = \Figwidth]{Frame 15 of Ambush\_3}{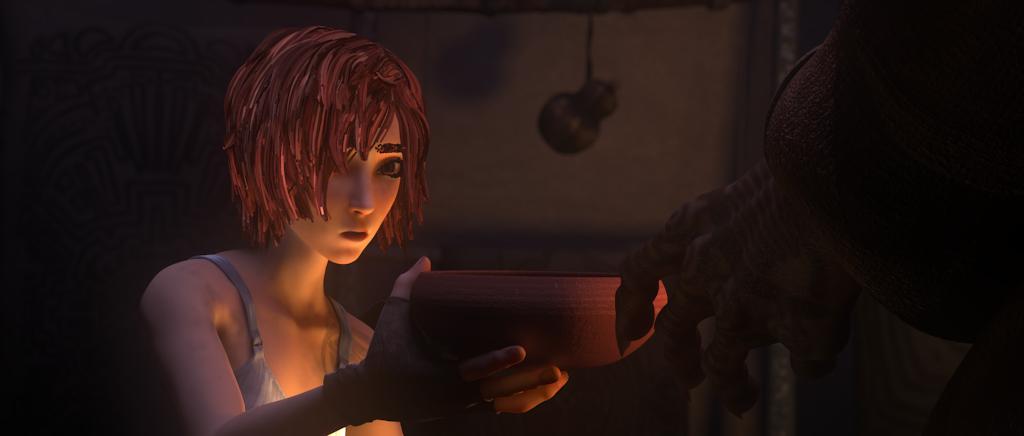} &
			\shiftfigure \hspace{-1em}\subfigimg[width = \Figwidth]{Frame 16}{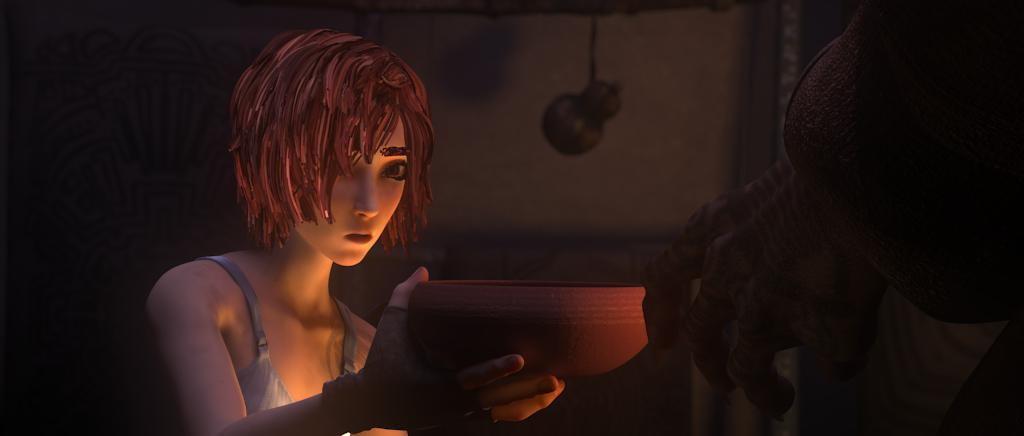} &
			\shiftfigure \hspace{-1em}\subfigimg[width = \Figwidth]{FlowNetC}{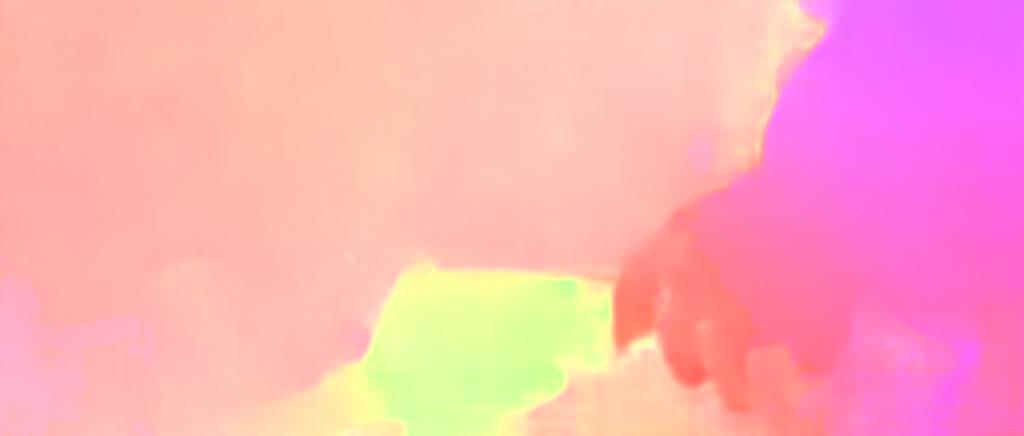} &
			\shiftfigure \hspace{-1em}\subfigimg[width = \Figwidth]{FlowNetC+}{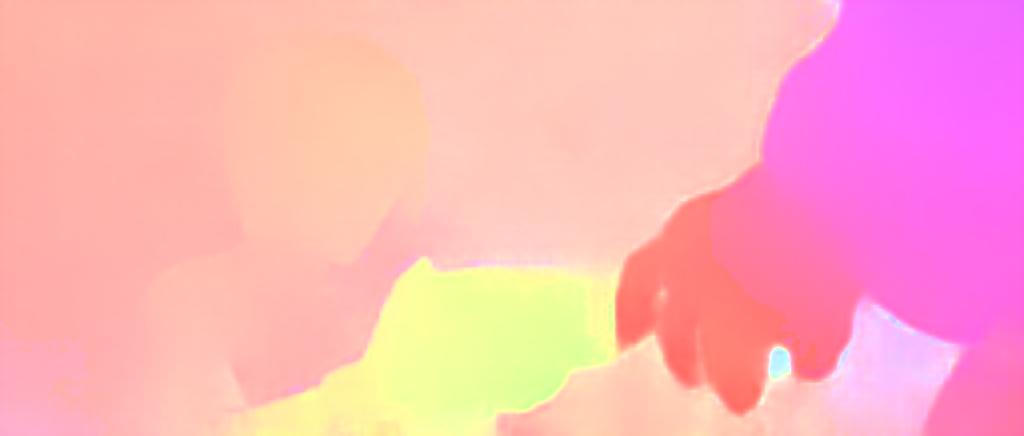} \\
			\shiftfigure \hspace{-1em}\subfigimg[width = \Figwidth]{Frame 11 of Market\_1}{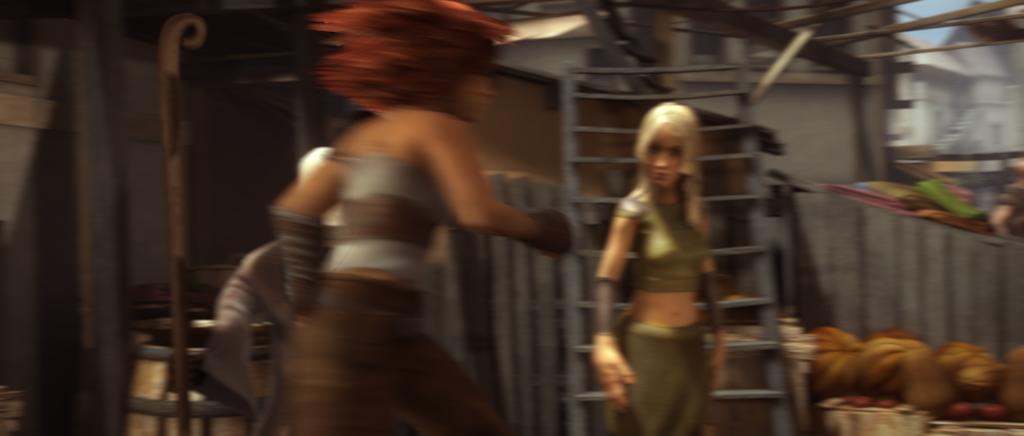} &	
			\shiftfigure \hspace{-1em}\subfigimg[width = \Figwidth]{Frame 12 }{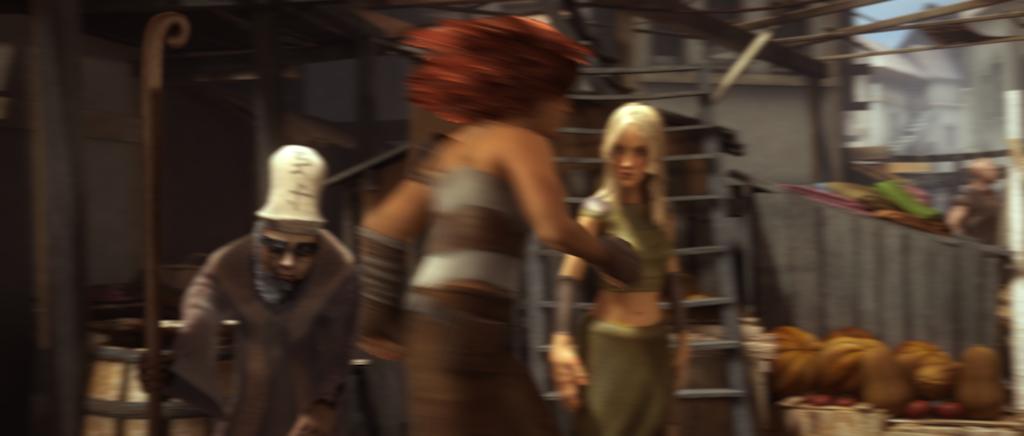} &	
			\shiftfigure \hspace{-1em}\subfigimg[width = \Figwidth]{FlowNetC}{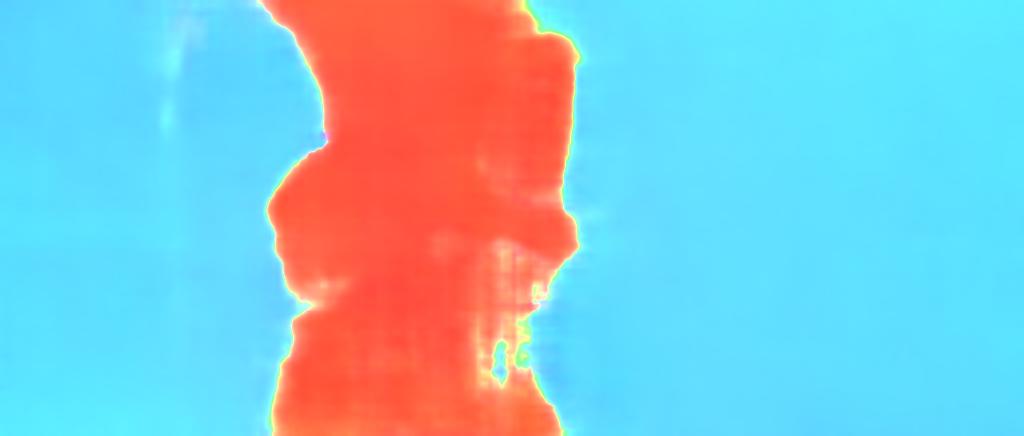} &				
			\shiftfigure \hspace{-1em}\subfigimg[width = \Figwidth]{FlowNetC+}{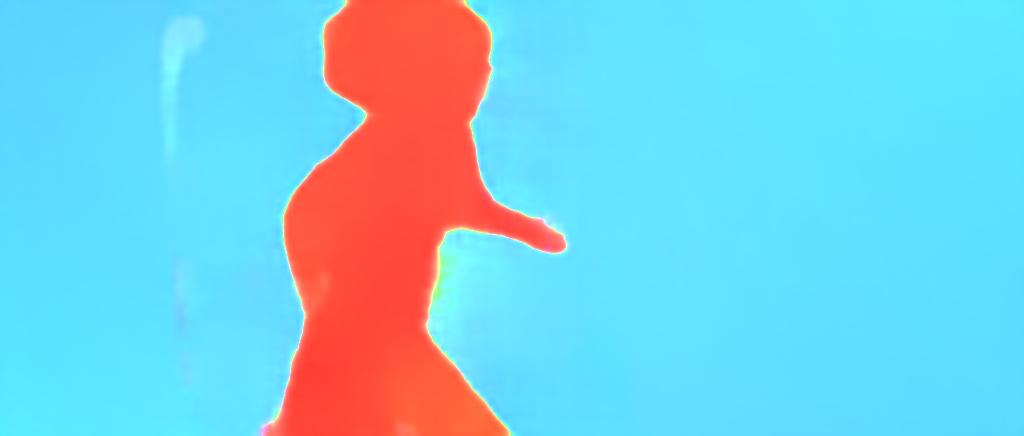} \\	
		\end{tabular}		
	\end{center}
	\vspace{-4mm}
	\caption{{\bf Training procedure matters}. FlowNetC and FlowNetC+ use the same network architecture but have been trained differently. FlowNetC+ has been trained using our procedure and generates results with finer details and  fewer artifacts than the previously trained FlowNetC.}
	\label{fig:flownet:comparison}
\end{figure*}

\subsection{Training Matters}
We used the same architecture in the first~\cite{Sun2017pwc} and second (CVPR) versions of our arXiv paper but observed an about 10\% improvement on the final pass of the Sintel test set. The performance improvement results solely from changes in the training procedures, including performing horizontal flips, not adding additive Gaussian noise, and disrupting the learning rate. One questions arises: how do other models perform using the same training procedure as PWC-Net?

To better understand the effects of models and training and fairly compare with existing methods, we re-train the FlowNetS and FlowNetC models using exactly the same training procedure as PWC-Net, including the robust training loss function. We name the retrained model as FlowNetS+ and FlowNetC+ respectively and evaluate them using the test set of Sintel, as summarized in Table~\ref{tab:result:sintel}. 
Figure~\ref{fig:flownet:comparison} shows the visual results FlowNetC trained using different training protocols. The results by FlowNetC+ have fewer artifacts and are more piece-wise smooth than the reviously trained FlowNetC. 
As shown in Table~\ref{tab:result:sintel}, FlowNetC+ is about 8\% less accurate on Sintel final and 3 times larger in model size than PWC-Net , which demonstrates the benefit of the new network architecture under the same training procedure.

To our surprise, FlowNetC+ is about 5\% more accurate than the published FlowNet2 model on the final pass, because FlowNet2 uses FlowNetC as a sub-network. We should note that this is not a fair comparison for FlowNet2, because we are unable to apply the same training protocol to the FlowNet2 model, which requires sequential training of several sub-networks. It is expected that a more careful, customized training schemes would improve the performance of FlowNet2. 

It is often assumed or taken for granted that the results published by authors represent the best possible performance of a method. However, our results show that we should carefully evaluate published methods to ``identify the source of empirical gains''~\cite{Lipton2018Troubling}. When we observe improvements over previous models, it is critical to analyze whether the gain results from the model or the training procedure. It would be informative to evaluate models trained in the same way or compare training procedures using the same model. 
To enable fair comparisons and further innovations, we will make our training protocols available. 
	
	\section{Comparison With Closely-related Work}	
	As the field is changing rapidly, it is informative to do a detailed discussion of the closely-related work. Both FlowNet2~\cite{Ilg:2016:Flownet2} and SpyNet~\cite{Ranjan:2016:SpyNet} have been designed using principles from stereo and optical flow. However, the architecture of PWC-Net has significant differences. 
	
	SpyNet uses image pyramids while PWC-Net learns feature pyramids. FlowNet2 uses three-level feature pyramids in the first module of its whole network, \ie, FlowNetC. By contrast, PWC-Net uses much deeper feature pyramids. As analyzed in the ablation study, using deeper feature pyramids usually leads to better performance. Both SpyNet and FlowNet2 warp the input images, while PWC-Net warps the features, which enables the information to propagate throughout the whole feature pyramids. 
	
	SpyNet feeds CNNs with images, while PWC-Net feeds a cost volume. As the cost volume is a more discriminative representation of the search space for optical flow, the learning task for CNNs becomes easier. FlowNet2/FlowNetC constructs the cost volume at a single resolution with a large search range. However, using features at a fixed resolution may not be effective at resolving the well-known ``aperture problem''~\cite{adelson1982phenomenal,Horn:1981:DO,Lucas:1981:LK,weiss2002motion}. By contrast, PWC-Net constructs multi-resolution cost volume and reduces the computation using a small search range.
	
	Regarding performance, PWC-Net outperforms SpyNet by a significant margin. Additionally, SpyNet has been trained sequentially, while PWC-Net can be trained end-to-end from scratch.	FlowNet2 achieves impressive performance by stacking several basic models into a large-capacity model. 
	The much smaller PWC-Net obtains similar or better performance by embedding classical principles into the network architecture. 
	It would be interesting to use PWC-Net as a building block to design large networks.
	
	Two recent papers also incorporate domain knowledge of flow into the CNN architectures. LiteFlowNet~\cite{Hui_2018_CVPR} uses similar ideas as PWC-Net, including feature pyramids, warping features, and constructing a cost volume with a limited search range at multiple resolutions. LiteFlowNet furthers incorporates a flow regularization layer to deal with outliers using a feature-driven local convolutions. However, LiteFlowNet requires sequential (stage-wise) training. 
	The CVPR final version of LiteFlowNet  (March. 2018) is about 8\% less accurate on Sintel final than the first arXiv version of PWC-Net~\cite{Sun2017pwc} published in Sep. 2017 (avg. EPE 6.09 vs. 5.63).  In an updated arXiv version~\cite{Hui_2018_arXiv} published in May 2018, LiteFlowNet uses similar data augmentation schemes as the CVPR final version of PWC-Net, \eg, not adding Gaussian noise, horizontal flipping (image mirroring), and reducing the spatial data augmentation for KITTI.  With these changes, LiteFlowNet reports performance close to PWC-Net on Sintel final (avg. EPE: 5.33 vs 5.04) and KITTI 2015 (F-all: 9.38\% vs. 9.60\%). This further confirms the importance of  training in obtaining top performance. 

	Another paper, TVNet~\cite{Fan_2018_CVPR}, subsumes a specific optical flow solver, the TV-L1 method~\cite{zach2007duality}, and is initialized by unfolding its optimization iterations as neural layers.  TVNet is used to learn rich and task-specific patterns and obtains excellent performance on activity classification. 
	The readers are urged to read these papers to better understand the similarities and differences.
	
	\begin{table}[h]
		\caption{ \bd{Comparison of network architectures.}  }
		\label{tab:model:comparison} 
		\centering
		\begin{tabular}{lccccccc} \\
			Principles &  FlowNetC & FlowNet2 & SpyNet  & PWC-Net  \\ \hline
			Pyramid &  {3-level} &{3-level} & Image & 6-level \\
			Warping & - & Image & Image & Feature \\
			\multirow{2}*{Cost volume} & {single-level} & {single-level}  & \multirow{2}*{-}  & {multi-level}\\
			 & {large range}& {large range}&  & {small range}\\
		\end{tabular}
	\end{table}

	\section{Conclusions}
	We have developed a compact but effective CNN model for optical flow estimation using simple and well-established principles: pyramidal processing, warping, and cost volume processing.
	Combining deep learning with domain knowledge not only reduces the model size but also improves the performance. 
	PWC-Net is about 17 times smaller in size, 2 times faster in inference, easier to train, and 11\% more accurate on Sintel final than the recent FlowNet2 model.
	It performs robustly across four different benchmarks using the same set of parameters and is the winning entry in the optical flow competition of the robust vision challenge. 

	We have also shown that the performance gains of PWC-Net result from both the new model architecture and the training procedures. 
	Retrained using our procedures, FlowNetC is even 5\% more accurate on Sintel final than the published FlowNet2, which uses FlowNetC as a sub-network. 
	We have further improved the training procedures, which increase the accuracy of PWC-Net on Sintel by 10\% and on KITTI 2012 and 2015 by 20\%. 
	The results show the complicated interplay between models and training and call for careful experimental designs to identify the sources of empirical gains. 
    To enable comparison and further innovations, we will make the retrained models and training protocols available  on \href{https://github.com/NVlabs/PWC-Net}{https://github.com/NVlabs/PWC-Net}.  
	
	\beforePara
	{
	\paragraph{Acknowledgements} We would like to thank Zhile Ren and Jinwei Gu for porting the Caffe code to PyTorch, Eddy Ilg for clarifying details about the FlowNet2 paper, Ming-Hsuan Yang for helpful suggestions, Michael Pellauer for proofreading,  github users for clarifying questions, and the anonymous reviewers at ICCV'17 and CVPR'18 for constructive comments.
	}
	
\bibliographystyle{IEEEtran}
\bibliography{ml_flow}

\begin{thebibliography}{10}
\providecommand{\url}[1]{#1}
\csname url@samestyle\endcsname
\providecommand{\newblock}{\relax}
\providecommand{\bibinfo}[2]{#2}
\providecommand{\BIBentrySTDinterwordspacing}{\spaceskip=0pt\relax}
\providecommand{\BIBentryALTinterwordstretchfactor}{4}
\providecommand{\BIBentryALTinterwordspacing}{\spaceskip=\fontdimen2\font plus
\BIBentryALTinterwordstretchfactor\fontdimen3\font minus
  \fontdimen4\font\relax}
\providecommand{\BIBforeignlanguage}[2]{{%
\expandafter\ifx\csname l@#1\endcsname\relax
\typeout{** WARNING: IEEEtran.bst: No hyphenation pattern has been}%
\typeout{** loaded for the language `#1'. Using the pattern for}%
\typeout{** the default language instead.}%
\else
\language=\csname l@#1\endcsname
\fi
#2}}
\providecommand{\BIBdecl}{\relax}
\BIBdecl

\bibitem{krizhevsky2012imagenet}
A.~Krizhevsky, I.~Sutskever, and G.~E. Hinton, ``{ImageNet} classification with
  deep convolutional neural networks,'' in \emph{Advances in Neural Information
  Processing Systems (NIPS)}, 2012.

\bibitem{russakovsky2015imagenet}
O.~Russakovsky, J.~Deng, H.~Su, J.~Krause, S.~Satheesh, S.~Ma, Z.~Huang,
  A.~Karpathy, A.~Khosla, M.~Bernstein \emph{et~al.}, ``Imagenet large scale
  visual recognition challenge,'' \emph{International Journal of Computer
  Vision (IJCV)}, vol. 115, no.~3, pp. 211--252, 2015.

\bibitem{Lecun1989Backpropagation}
Y.~LeCun, B.~Boser, J.~S. Denker, D.~Henderson, R.~E. Howard, W.~Hubbard, and
  L.~D. Jackel, ``Backpropagation applied to handwritten zip code
  recognition,'' \emph{Neural computation}, 1989.

\bibitem{simonyan2014very}
K.~Simonyan and A.~Zisserman, ``Very deep convolutional networks for
  large-scale image recognition,'' \emph{arXiv preprint arXiv:1409.1556}, 2014.

\bibitem{szegedy2015going}
C.~Szegedy, W.~Liu, Y.~Jia, P.~Sermanet, S.~Reed, D.~Anguelov, D.~Erhan,
  V.~Vanhoucke, and A.~Rabinovich, ``Going deeper with convolutions,'' in
  \emph{IEEE Conference on Computer Vision and Pattern Recognition (CVPR)},
  2015.

\bibitem{He2016Deep}
K.~He, X.~Zhang, S.~Ren, and J.~Sun, ``Deep residual learning for image
  recognition,'' in \emph{IEEE Conference on Computer Vision and Pattern
  Recognition (CVPR)}, 2016.

\bibitem{huang2016densely}
G.~Huang, Z.~Liu, K.~Q. Weinberger, and L.~van~der Maaten, ``Densely connected
  convolutional networks,'' in \emph{IEEE Conference on Computer Vision and
  Pattern Recognition (CVPR)}, 2017.

\bibitem{Dosovitskiy:2015Flownet}
A.~Dosovitskiy, P.~Fischery, E.~Ilg, C.~Hazirbas, V.~Golkov, P.~van~der Smagt,
  D.~Cremers, T.~Brox \emph{et~al.}, ``{FlowNet}: Learning optical flow with
  convolutional networks,'' in \emph{IEEE International Conference on Computer
  Vision (ICCV)}, 2015.

\bibitem{ronneberger2015u}
O.~Ronneberger, P.~Fischer, and T.~Brox, ``{U-Net}: Convolutional networks for
  biomedical image segmentation,'' in \emph{International Conference on Medical
  Image Computing and Computer Assisted Intervention (MICCAI)}, 2015.

\bibitem{Ilg:2016:Flownet2}
E.~Ilg, N.~Mayer, T.~Saikia, M.~Keuper, A.~Dosovitskiy, and T.~Brox,
  ``{FlowNet} 2.0: Evolution of optical flow estimation with deep networks,''
  in \emph{IEEE Conference on Computer Vision and Pattern Recognition (CVPR)},
  2017.

\bibitem{Ranjan:2016:SpyNet}
A.~Ranjan and M.~J. Black, ``Optical flow estimation using a spatial pyramid
  network,'' in \emph{IEEE Conference on Computer Vision and Pattern
  Recognition (CVPR)}, 2017.

\bibitem{Baker:2011:DEO}
S.~Baker, D.~Scharstein, J.~P. Lewis, S.~Roth, M.~J. Black, and R.~Szeliski,
  ``A database and evaluation methodology for optical flow,''
  \emph{International Journal of Computer Vision (IJCV)}, 2011.

\bibitem{weber1995robust}
J.~Weber and J.~Malik, ``Robust computation of optical flow in a multi-scale
  differential framework,'' \emph{International Journal of Computer Vision
  (IJCV)}, vol.~14, no.~1, pp. 67--81, 1995.

\bibitem{Hosni2013Fast}
A.~Hosni, C.~Rhemann, M.~Bleyer, C.~Rother, and M.~Gelautz, ``Fast cost-volume
  filtering for visual correspondence and beyond,'' \emph{IEEE Transactions on
  Pattern Analysis and Machine Intelligence (TPAMI)}, 2013.

\bibitem{Scharstein:2002:TEDS}
D.~Scharstein and R.~Szeliski, ``A taxonomy and evaluation of dense two-frame
  stereo correspondence algorithms,'' \emph{International Journal of Computer
  Vision (IJCV)}, 2002.

\bibitem{Zbontar2016Stereo}
J.~Zbontar and Y.~LeCun, ``Stereo matching by training a convolutional neural
  network to compare image patches,'' \emph{Journal of Machine Learning
  Research (JMLR)}, 2016.

\bibitem{Xu2017Accurate}
J.~Xu, R.~Ranftl, and V.~Koltun, ``Accurate optical flow via direct cost volume
  processing,'' in \emph{IEEE Conference on Computer Vision and Pattern
  Recognition (CVPR)}, 2017.

\bibitem{Lecun1998Gradient}
Y.~LeCun, L.~Bottou, Y.~Bengio, and P.~Haffner, ``Gradient-based learning
  applied to document recognition,'' \emph{Proceedings of the IEEE}, vol.~86,
  no.~11, pp. 2278--2324, 1998.

\bibitem{Behl2017ICCV}
A.~Behl, O.~H. Jafari, S.~K. Mustikovela, H.~A. Alhaija, C.~Rother, and
  A.~Geiger, ``Bounding boxes, segmentations and object coordinates: How
  important is recognition for {3D} scene flow estimation in autonomous driving
  scenarios?'' in \emph{IEEE International Conference on Computer Vision
  (ICCV)}, 2017.

\bibitem{Horn:1981:DO}
B.~Horn and B.~Schunck, ``Determining optical flow,'' \emph{Artificial
  Intelligence}, 1981.

\bibitem{Black:1996:REMO}
M.~J. Black and P.~Anandan, ``The robust estimation of multiple motions:
  Parametric and piecewise-smooth flow fields,'' \emph{Computer Vision and
  Image Understanding (CVIU)}, 1996.

\bibitem{Bruhn:2005:CLG}
A.~Bruhn, J.~Weickert, and C.~Schn{\"o}rr, ``Lucas/{K}anade meets
  {H}orn/{S}chunck: combining local and global optic flow methods,''
  \emph{International Journal of Computer Vision (IJCV)}, 2005.

\bibitem{Brox:2004:HAO}
T.~Brox, A.~Bruhn, N.~Papenberg, and J.~Weickert, ``High accuracy optical flow
  estimation based on a theory for warping,'' in \emph{European Conference on
  Computer Vision (ECCV)}, 2004.

\bibitem{Sun:IJCV:2014}
D.~Sun, S.~Roth, and M.~J. Black, ``A quantitative analysis of current
  practices in optical flow estimation and the principles behind them,''
  \emph{International Journal of Computer Vision (IJCV)}, 2014.

\bibitem{Brox:LDOF:2011}
T.~Brox and J.~Malik, ``Large displacement optical flow: Descriptor matching in
  variational motion estimation,'' \emph{IEEE Transactions on Pattern Analysis
  and Machine Intelligence (TPAMI)}, 2011.

\bibitem{Weinzaepfel:2013:DeepFlow}
P.~Weinzaepfel, J.~Revaud, Z.~Harchaoui, and C.~Schmid, ``{DeepFlow: Large
  displacement optical flow with deep matching},'' in \emph{IEEE International
  Conference on Computer Vision (ICCV)}, 2013.

\bibitem{Xu:2012:MDP}
L.~Xu, J.~Jia, and Y.~Matsushita, ``Motion detail preserving optical flow
  estimation,'' \emph{IEEE Transactions on Pattern Analysis and Machine
  Intelligence (TPAMI)}, 2012.

\bibitem{EpicFlow}
J.~Revaud, P.~Weinzaepfel, Z.~Harchaoui, and C.~Schmid, ``{EpicFlow}:
  Edge-preserving interpolation of correspondences for optical flow,'' in
  \emph{IEEE Conference on Computer Vision and Pattern Recognition (CVPR)},
  2015.

\bibitem{Bai2016Exploiting}
M.~Bai, W.~Luo, K.~Kundu, and R.~Urtasun, ``Exploiting semantic information and
  deep matching for optical flow,'' in \emph{European Conference on Computer
  Vision (ECCV)}, 2016.

\bibitem{Bailer_2017_CVPR}
C.~Bailer, K.~Varanasi, and D.~Stricker, ``{CNN}-based patch matching for
  optical flow with thresholded hinge embedding loss,'' in \emph{IEEE
  Conference on Computer Vision and Pattern Recognition (CVPR)}, 2017.

\bibitem{Chen2016Full}
Q.~Chen and V.~Koltun, ``Full flow: Optical flow estimation by global
  optimization over regular grids,'' in \emph{IEEE Conference on Computer
  Vision and Pattern Recognition (CVPR)}, 2016.

\bibitem{Hu2016Efficient}
Y.~Hu, R.~Song, and Y.~Li, ``Efficient coarse-to-fine patchmatch for large
  displacement optical flow,'' in \emph{IEEE Conference on Computer Vision and
  Pattern Recognition (CVPR)}, 2016.

\bibitem{yang2017s2f}
Y.~Yang and S.~Soatto, ``S2f: Slow-to-fast interpolator flow,'' in \emph{IEEE
  Conference on Computer Vision and Pattern Recognition (CVPR)}, 2017.

\bibitem{Zweig_2017_CVPR}
S.~Zweig and L.~Wolf, ``Interponet, a brain inspired neural network for optical
  flow dense interpolation,'' in \emph{IEEE Conference on Computer Vision and
  Pattern Recognition (CVPR)}, 2017.

\bibitem{Wulff2017Optical}
J.~Wulff, L.~Sevilla-Lara, and M.~J. Black, ``Optical flow in mostly rigid
  scenes,'' in \emph{IEEE Conference on Computer Vision and Pattern Recognition
  (CVPR)}, 2017.

\bibitem{Simoncelli:1991:Probability}
E.~P. Simoncelli, E.~H. Adelson, and D.~J. Heeger, ``Probability distributions
  of optical flow,'' in \emph{IEEE Conference on Computer Vision and Pattern
  Recognition (CVPR)}, 1991.

\bibitem{Freeman:2000:LLLV}
W.~T. Freeman, E.~C. Pasztor, and O.~T. Carmichael, ``Learning low-level
  vision,'' \emph{International Journal of Computer Vision (IJCV)}, 2000.

\bibitem{Roth:2007:SSO}
S.~Roth and M.~J. Black, ``On the spatial statistics of optical flow,''
  \emph{International Journal of Computer Vision (IJCV)}, 2007.

\bibitem{Sun:2008:LOF}
D.~Sun, S.~Roth, J.~P. Lewis, and M.~J. Black, ``Learning optical flow,'' in
  \emph{European Conference on Computer Vision (ECCV)}, 2008.

\bibitem{Li2008Learning}
Y.~Li and D.~P. Huttenlocher, ``Learning for optical flow using stochastic
  optimization,'' in \emph{European Conference on Computer Vision (ECCV)},
  2008.

\bibitem{Wulff:2015:PCA}
J.~Wulff and M.~J. Black, ``Efficient sparse-to-dense optical flow estimation
  using a learned basis and layers,'' in \emph{IEEE Conference on Computer
  Vision and Pattern Recognition (CVPR)}, 2015, pp. 120--130.

\bibitem{Werlberger:2009:AHOF}
M.~Werlberger, W.~Trobin, T.~Pock, A.~Wedel, D.~Cremers, and H.~Bischof,
  ``Anisotropic {H}uber-{L1} optical flow,'' in \emph{British Machine Vision
  Conference (BMVC)}, 2009.

\bibitem{Mayer:2016:Large}
N.~Mayer, E.~Ilg, P.~H{\"a}usser, P.~Fischer, D.~Cremers, A.~Dosovitskiy, and
  T.~Brox, ``A large dataset to train convolutional networks for disparity,
  optical flow, and scene flow estimation,'' in \emph{IEEE Conference on
  Computer Vision and Pattern Recognition (CVPR)}, 2016.

\bibitem{Memisevic:2007:Unsupervised}
R.~Memisevic and G.~Hinton, ``Unsupervised learning of image transformations,''
  in \emph{IEEE Conference on Computer Vision and Pattern Recognition (CVPR)},
  2007.

\bibitem{Long:2016:Learning}
G.~Long, L.~Kneip, J.~M. Alvarez, H.~Li, X.~Zhang, and Q.~Yu, ``Learning image
  matching by simply watching video,'' in \emph{European Conference on Computer
  Vision (ECCV)}, 2016.

\bibitem{YuHD16}
J.~J. Yu, A.~W. Harley, and K.~G. Derpanis, ``Back to basics: Unsupervised
  learning of optical flow via brightness constancy and motion smoothness,'' in
  \emph{CoRR}, 2016.

\bibitem{Lai2017SemiFlowGAN}
W.-S. Lai, J.-B. Huang, and M.-H. Yang, ``Semi-supervised learning for optical
  flow with generative adversarial networks,'' in \emph{Advances in Neural
  Information Processing Systems (NIPS)}, 2017.

\bibitem{Barron:1994:PO}
J.~Barron, D.~Fleet, and S.~Beauchemin, ``Performance of optical flow
  techniques,'' \emph{International Journal of Computer Vision (IJCV)}, 1994.

\bibitem{Liu:2008:HAMA}
C.~Liu, W.~T. Freeman, E.~H. Adelson, and Y.~Weiss, ``Human-assisted motion
  annotation,'' in \emph{IEEE Conference on Computer Vision and Pattern
  Recognition (CVPR)}, 2008.

\bibitem{Geiger:2012:KITTI}
A.~Geiger, P.~Lenz, and R.~Urtasun, ``Are we ready for autonomous driving?
  {T}he {KITTI} vision benchmark suite,'' in \emph{IEEE Conference on Computer
  Vision and Pattern Recognition (CVPR)}, 2012.

\bibitem{Menze2015CVPR}
M.~Menze and A.~Geiger, ``Object scene flow for autonomous vehicles,'' in
  \emph{IEEE Conference on Computer Vision and Pattern Recognition (CVPR)},
  2015.

\bibitem{Butler:ECCV:2012}
D.~J. Butler, J.~Wulff, G.~B. Stanley, and M.~J. Black, ``A naturalistic open
  source movie for optical flow evaluation,'' in \emph{European Conference on
  Computer Vision (ECCV)}, 2012.

\bibitem{vincent2008extracting}
P.~Vincent, H.~Larochelle, Y.~Bengio, and P.-A. Manzagol, ``Extracting and
  composing robust features with denoising autoencoders,'' in
  \emph{International Conference on Machine Learning (ICML)}, 2008.

\bibitem{chen2017deeplab}
L.~C. Chen, G.~Papandreou, I.~Kokkinos, K.~Murphy, and A.~L. Yuille,
  ``{DeepLab}: Semantic image segmentation with deep convolutional nets, atrous
  convolution, and fully connected {CRFs},'' \emph{IEEE Transactions on Pattern
  Analysis and Machine Intelligence (TPAMI)}, 2017.

\bibitem{yu2015multi}
F.~Yu and V.~Koltun, ``Multi-scale context aggregation by dilated
  convolutions,'' in \emph{International Conference on Learning Representations
  (ICLR)}, 2016.

\bibitem{Jegou:2016:Densenet}
S.~J{\'e}gou, M.~Drozdzal, D.~Vazquez, A.~Romero, and Y.~Bengio, ``The one
  hundred layers tiramisu: Fully convolutional densenets for semantic
  segmentation,'' in \emph{IEEE Conference on Computer Vision and Pattern
  Recognition (CVPR) Workshop}, 2017.

\bibitem{Wedel:2008:ITVL1}
A.~Wedel, T.~Pock, C.~Zach, D.~Cremers, and H.~Bischof, ``An improved algorithm
  for {TV-L1} optical flow,'' in \emph{Dagstuhl Motion Workshop}, 2008.

\bibitem{Xiao:2006:BFO}
J.~Xiao, H.~Cheng, H.~Sawhney, C.~Rao, and M.~Isnardi, ``Bilateral
  filtering-based optical flow estimation with occlusion detection,'' in
  \emph{European Conference on Computer Vision (ECCV)}, 2006.

\bibitem{Jaderberg:2015:Spatial}
M.~Jaderberg, K.~Simonyan, A.~Zisserman \emph{et~al.}, ``Spatial transformer
  networks,'' in \emph{Advances in Neural Information Processing Systems
  (NIPS)}, 2015.

\bibitem{jia2014caffe}
Y.~Jia, E.~Shelhamer, J.~Donahue, S.~Karayev, J.~Long, R.~Girshick,
  S.~Guadarrama, and T.~Darrell, ``Caffe: Convolutional architecture for fast
  feature embedding,'' in \emph{ACM Multimedia}, 2014.

\bibitem{Gadot2016PatchBatch}
D.~Gadot and L.~Wolf, ``{PatchBatch}: A batch augmented loss for optical
  flow,'' in \emph{IEEE Conference on Computer Vision and Pattern Recognition
  (CVPR)}, 2016.

\bibitem{Bailer2015Flow}
C.~Bailer, B.~Taetz, and D.~Stricker, ``Flow fields: Dense correspondence
  fields for highly accurate large displacement optical flow estimation,'' in
  \emph{IEEE International Conference on Computer Vision (ICCV)}, 2015.

\bibitem{Hur_2017_ICCV}
J.~Hur and S.~Roth, ``{MirrorFlow}: Exploiting symmetries in joint optical flow
  and occlusion estimation,'' in \emph{IEEE International Conference on
  Computer Vision (ICCV)}, Oct 2017.

\bibitem{ren2017cascaded}
Z.~Ren, D.~Sun, J.~Kautz, and E.~Sudderth, ``Cascaded scene flow prediction
  using semantic segmentation,'' in \emph{3DV}, 2017.

\bibitem{kondermann2016hci}
D.~Kondermann, R.~Nair, K.~Honauer, K.~Krispin, J.~Andrulis, A.~Brock,
  B.~Gussefeld, M.~Rahimimoghaddam, S.~Hofmann, C.~Brenner \emph{et~al.}, ``The
  hci benchmark suite: Stereo and flow ground truth with uncertainties for
  urban autonomous driving,'' in \emph{CVPR Workshops}, 2016, pp. 19--28.

\bibitem{Sun2017pwc}
D.~Sun, X.~Yang, M.-Y. Liu, and J.~Kautz, ``{PWC-Net}: {CNNs} for optical flow
  using pyramid, warping, and cost volume,'' \emph{arXiv preprint
  arXiv:1709.02371}, 2017.

\bibitem{Lipton2018Troubling}
Z.~C. Lipton and J.~Steinhardt, ``Troubling trends in machine learning
  scholarship,'' \emph{arXiv preprint arXiv:1807.03341}, 2018.

\bibitem{adelson1982phenomenal}
E.~H. Adelson and J.~A. Movshon, ``Phenomenal coherence of moving visual
  patterns,'' \emph{Nature}, vol. 300, no. 5892, p. 523, 1982.

\bibitem{Lucas:1981:LK}
B.~Lucas and T.~Kanade, ``An iterative image registration technique with an
  application to stereo vision,'' 1981, pp. 674--679.

\bibitem{weiss2002motion}
Y.~Weiss, E.~P. Simoncelli, and E.~H. Adelson, ``Motion illusions as optimal
  percepts,'' \emph{Nature neuroscience}, vol.~5, no.~6, p. 598, 2002.

\bibitem{Hui_2018_CVPR}
T.-W. Hui, X.~Tang, and C.~Change~Loy, ``Liteflownet: {A} lightweight
  convolutional neural network for optical flow estimation,'' in \emph{IEEE
  Conference on Computer Vision and Pattern Recognition (CVPR)}, 2018.

\bibitem{Hui_2018_arXiv}
T.~Hui, X.~Tang, and C.~C. Loy, ``Liteflownet: {A} lightweight convolutional
  neural network for optical flow estimation,'' \emph{arXiv preprint
  arXiv:1805.07036}, 2018.

\bibitem{Fan_2018_CVPR}
L.~Fan, W.~Huang, C.~Gan, S.~Ermon, B.~Gong, and J.~Huang, ``End-to-end
  learning of motion representation for video understanding,'' in \emph{IEEE
  Conference on Computer Vision and Pattern Recognition (CVPR)}, 2018.

\bibitem{zach2007duality}
C.~Zach, T.~Pock, and H.~Bischof, ``A duality based approach for realtime tv-l
  1 optical flow,'' in \emph{German Conference on Pattern Recognition (DAGM)},
  2007.

\end{thebibliography}

\begin{figure}[h]
	\begin{center}
		\newcommand{\shiftfigure}{\hspace{5mm}}
		\begin{tabular}{cc}
			\includegraphics[width=0.45\linewidth]{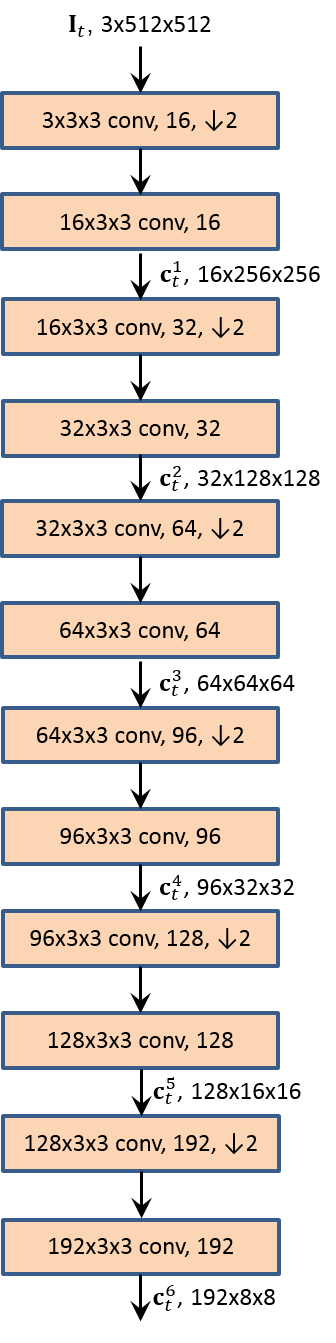} 
		\end{tabular}
	\end{center}
	\vspace{-4mm}
	\caption{The feature pyramid extractor network. The first image ($t\!=\!1$) and the second image ($t\!=\!2$) are encoded using the same Siamese network. Each convolution is followed by a leaky ReLU unit. The convolutional layer and the $\times2$ downsampling layer at each level is implemented using a single convolutional layer with a stride of 2. $\mv{c}^l_t$ denotes  extracted features of image $t$ at level $l$; }
	\label{fig:network:encoder}
\end{figure}

\begin{figure}
	\begin{center}
		\newcommand{\shiftfigure}{\hspace{5mm}}
		\begin{tabular}{cc}
			\includegraphics[width=0.8\linewidth]{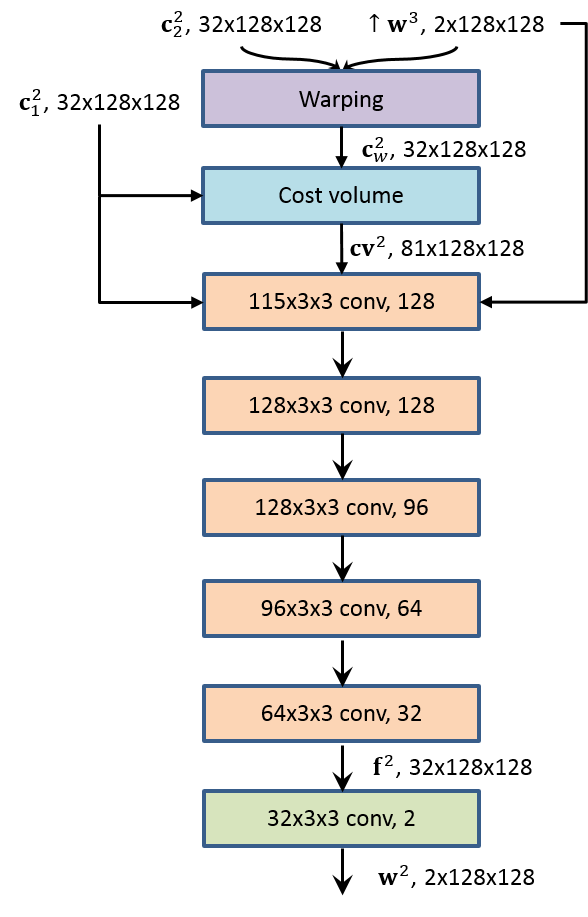} 
		\end{tabular}
	\end{center}
	\vspace{-4mm}
	\caption{The optical flow estimator network at pyramid level 2. Each convolutional layer is followed by a leaky ReLU unit except the last (light green) one that outputs the optical flow. }
	\label{fig:network:decoder2}
\end{figure}

\begin{figure}
	\begin{center}
		\newcommand{\shiftfigure}{\hspace{5mm}}
		\begin{tabular}{cc}
			\includegraphics[width=0.65\linewidth]{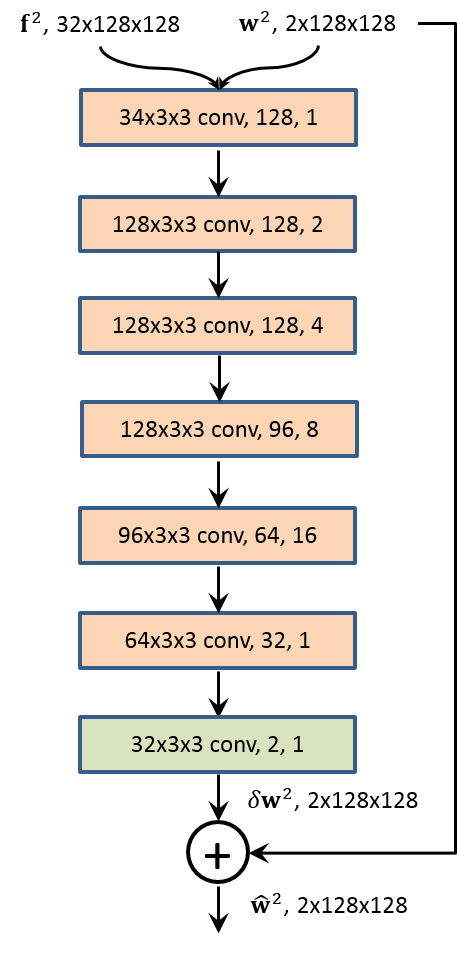} 
		\end{tabular}
	\end{center}
	\vspace{-4mm}
	\caption{The context network at pyramid level 2. Each convolutional layer is followed by a leaky ReLU unit except the last (light green) one that outputs the optical flow. The last number in each convolutional layer denotes the dilation constant. }
	\label{fig:network:pp}
\end{figure}

\appendix
In this appendix, we proivde more details about PWC-Net.
Figure~\ref{fig:network:encoder} shows the architecture for the 7-level feature pyramid extractor network used in our experiment. Note that the bottom level consists of the original input images. 
Figure~\ref{fig:network:decoder2} shows the optical flow estimator network at pyramid level 2. The optical flow estimator networks at other levels have the same structure except for the top level, which does not have the upsampled optical flow and directly computes cost volume using features of the first and second images. Figure~\ref{fig:network:pp} shows the context network that is adopted only at pyramid level 2. 

\end{document}